\documentclass[aps,pra,reprint,amsmath,amssymb,superscriptaddress,onecolumn,longbibliography,nofootinbib,notitlepage]{revtex4-2}
\usepackage{mathtools}
\usepackage{siunitx}
\usepackage[hidelinks]{hyperref}
\usepackage{svg}
\usepackage{caption}
\usepackage{float}
\usepackage{subcaption}
\usepackage{amsmath}
\usepackage[utf8]{inputenc}
\usepackage[braket, qm]{qcircuit}
\usepackage{bbm}
\usepackage{comment}

\usepackage{multirow}
\usepackage{xcolor}
\usepackage{soul}
\newcommand{\hlnew}[1]{#1}
\newcommand{\hlred}[1]{#1}
\newcommand{\hlredbf}[1]{\textbf{#1}}
\def\fromplm#1{}
\def\fromlgw#1{}
\def\fromtw#1{}
\def\fromplmnew#1{}
\def\fromtwnew#1{}

\usepackage[most]{tcolorbox}
\usepackage{tikz}
\newtcolorbox{redbox}[1]{colback=red!5!white,colframe=red!75!black,fonttitle=\bfseries,title=#1}

\begin{document}
\title{Physical Foundation Models:\\Fixed hardware implementations of large-scale neural networks}
\author{Logan~G.~Wright}
\email{logan.wright@yale.edu}
\affiliation{Department of Applied Physics, Yale University, New Haven, CT 06520, USA}
\affiliation{School of Applied and Engineering Physics, Cornell University, Ithaca, NY 14853, USA}

\author{Tianyu~Wang}
\affiliation{Department of Electrical and Computer Engineering, Boston University, Boston, MA 02215, USA}
\affiliation{School of Applied and Engineering Physics, Cornell University, Ithaca, NY 14853, USA}

\author{Tatsuhiro~Onodera}
\affiliation{School of Applied and Engineering Physics, Cornell University, Ithaca, NY 14853, USA}
\affiliation{NTT Physics and Informatics Laboratories, NTT Research, Inc., Sunnyvale, CA 94085, USA}

\author{Peter~L.~McMahon}
\email{pmcmahon@cornell.edu}
\affiliation{School of Applied and Engineering Physics, Cornell University, Ithaca, NY 14853, USA}
\affiliation{Kavli Institute at Cornell for Nanoscale Science, Cornell University, Ithaca, NY 14853, USA}

\begin{abstract}
Foundation models are deep neural networks (such as GPT-5, Gemini~3, and Llama~4) that have been trained on large datasets and can perform diverse downstream tasks---such as text or code generation, question answering, summarization, image classification, and image retrieval. The philosophy of foundation models is to put effort into making a single, large (${\sim}10^{12}$-parameter) general-purpose model that can be adapted to perform many downstream tasks, typically with no or minimal additional training. In this perspective piece, we explain how the recent rise of foundation models provides an exciting opportunity for hardware engineers: in contrast to when different models were used for different tasks, it now makes sense to build special-purpose, fixed hardware implementations of neural networks (for example, where the neural-network parameters may be stored in read-only memory), which could be manufactured and released at roughly 1 year cadence at which new versions of foundation models like GPT, Gemini, and Claude have been released.

We advocate for not just conventional implementations of neural-network inference hardware using digital-electronic matrix-multiplication circuits with read-only memory, but for a more radical re-thinking of how to implement large-scale artificial intelligence (AI). We argue that exotic hardware implementations where the neural network is realized directly at the level of the physical design of the hardware, and operates based on the hardware's natural physical dynamics---\textit{Physical Foundation Models}---could enable orders-of-magnitude advantages in energy efficiency, speed, and parameter density. For models at the scale of ${\sim}10^{12}$ parameters, such improvements would both greatly reduce the already-high energy burden of AI in datacenters and enhance the capabilities of AI in edge devices, which are generally power-constrained to much smaller models currently. There is also the potential for PFMs to enable inference hardware capable of running models that are much larger than the current largest models: PFMs supporting $10^{15}$, or perhaps even $10^{18}$, parameters seem plausible by some measures. We present back-of-the-envelope calculations to illustrate the scaling potential of PFMs using an optical example---a 3D nanostructured glass medium---and discuss prospects for PFMs in nanoelectronics and other physical platforms. We conclude with a discussion of the major research challenges and open questions that must be resolved if this vision of trillion-parameter PFMs and beyond is to become a reality.

\end{abstract}

\maketitle

\section{Introduction}
\label{sec:intro}

The energy costs, as well as capital costs, of servicing over 1~billion users of artificial-intelligence (AI) products such as OpenAI ChatGPT and Google Gemini present major practical challenges. The AI workload of datacenters worldwide already demands 44~gigawatts of power, with a projected increase to $>$150~gigawatts by 2030, and an anticipated capital cost of $>$\$5~trillion for building sufficient datacenter capacity and related infrastructure to meet the growing demand for AI~\cite{noffsinger2025cost}. The energy cost of running state-of-the-art models is also sufficiently high that these models are beyond the power (and compute and memory) budget of edge devices~\cite{girija2025optimizing}. In this concept paper, we are motivated by addressing the compute and energy challenges posed both by current-scale models and by potentially far larger models. Our focus here is on challenges in AI \textit{inference}---as opposed to \textit{training}, although we do also discuss training where it relates to the research agenda for inference hardware we propose.

Driven by these pressures, several recent efforts have pushed the specialization of inference hardware to remarkable extremes. Groq's Language Processing Unit (LPU)~\cite{groq2022lpu,abts2022think} and Cerebras' Wafer-Scale Engine~\cite{lie2023cerebras} leverage advanced packaging and large amounts of on-chip memory to dramatically reduce data movement costs. Most recently, the startup Taalas~\cite{taalas} has unveiled HC1, a ``direct-to-silicon'' inference ASIC in which a specific large-language model (Llama~3.1~8B) is hard-wired into the chip at fabrication, with reported throughputs on the order of ${\sim}10^4$ tokens/second/user---roughly $10\times$ faster, and consuming roughly $10\times$ less power, than GPU-based inference of the same model. Even with these gains, hardwiring weights addresses only part of the inference cost: making the parameters read-only helps with weight-fetch costs, but it cannot eliminate the activation-related data movement in Transformer inference (e.g., the key-value caches and pairwise activation--activation products in Attention~\cite{vaswani2017attention} are not ``weights-in-place'' and so are not directly addressed by hardwiring weights). In other words, going to read-only memory only helps so much. Beyond Taalas, Liu~et~al.~\cite{liu2025hnlpu} proposed a Hardwired-Neurons LPU (HNLPU) that physically embeds the weight parameters of a large language model into the metal interconnect topology of a fabricated chip, and predicts a ${\sim}1{,}000\times$ energy-efficiency improvement relative to GPUs. Their metal-embedding methodology encodes weights in the 3D routing of metal wires, eliminating weight fetches entirely and bringing hardwired neural-network inference---an idea explored since the 1980s~\cite{graf1988vlsi}---into the realm of economic viability for modern-scale models.

In our paper, we push the idea of hardwired inference to its absolute physical and logical limit: rather than encoding weights in read-only memory or the routing topology of a \textit{digital circuit}, we consider \textit{analog physical media} (e.g., nanostructured optical and semiconductor materials) in which the hardware's natural physical dynamics directly perform the entire (or most of the) neural-network computation. Critically, this approach also goes beyond hardwiring weights alone: by performing the entire forward computation---including the analogs of nonlinear activations, activation--activation interactions, skip connections, etc.---directly in physical dynamics, PFMs can in principle address some, and perhaps even all, of the very data-movement challenges that hardwired-weight digital ASICs like Taalas's leave unsolved. This approach can thus push inference costs to their absolute physical limits, taking advantage of fundamental, physics-derived gains on the energy, speed, and scale of inference operations that no amount of digital circuit optimization can surpass.

\subsection{Motivation 1: Making current-scale ($\lesssim 10^{12}$-parameter) AI models more energy-efficient}

AI systems based on deep neural networks have been growing rapidly \cite{kaplan2020scaling}. Today's large AI models routinely have $>10^{11}$ parameters~\cite{chowdhery2023palm,dubey2024llama,liu2024deepseek,meta2025llama4}, which is a driving factor of the large compute, memory, and energy requirements to run them. Much has been written about the challenges facing both datacenter and edge deployment of current AI models from these compute, memory, and energy demands. If one could find a way to dramatically reduce the energy cost of AI while preserving the quality of its output, it should be possible to slash the power requirements for datacenters and improve output quality and battery life in edge scenarios. To the extent that power considerations already limit model size and output quality in cloud scenarios, dramatically improving energy efficiency could also rapidly enable improved output quality for AI delivered by datacenters.

\subsection{Motivation 2: Enabling ultra-large-scale ($\gtrsim 10^{15}$-parameter) AI models in the future}

The number of parameters reached in current AI models ($>10^{11}$) would have seemed far-fetched, if not comical, decades ago. Tomorrow's models might, provided there are economically feasible paths, reach scales that we would find comical today. Scaling to large model sizes is presently a trend in the development of AI systems, motivated by the emergence of qualitatively new capabilities with scale, and by the performance improvements predicted by neural scaling laws \cite{Hestness2017,kaplan2020scaling,Hoffmann2022,Bahri2024}. With that in mind, we ask here: how can we possibly construct AI systems that are much, much larger; models that have $10^{15}$, $10^{18}$, $10^{21}$, or even many, many more parameters?

As of 2026, the approximate default answer to this question is that AI systems will, whenever possible, run on digital supercomputers on the cloud, and applications of such AI systems will be limited by latency, uptime, and costs\footnote{Where costs in dollars, energy consumption, and carbon emissions may each play a role in constraining the commercially and societally viable scale and applications of AI.}. This is plausible, and for many applications of AI it may be adequate. But is it the only way? And it is the best way?

\begin{figure}
    \centering
    \includegraphics[width=\linewidth]{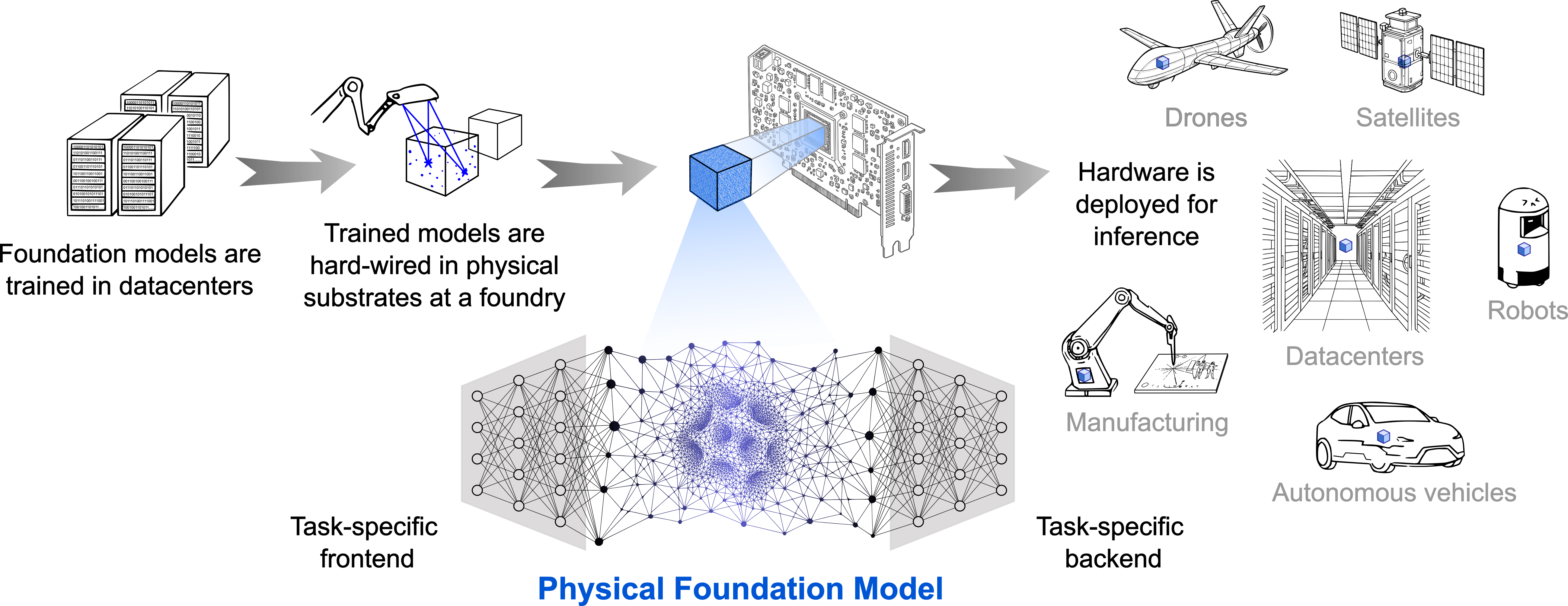}
    \caption{\textbf{Physical Foundation Models}: PFMs are fixed (hard-wired) pieces of physical hardware that execute neural-network \emph{inference}. Because they are fixed at the time of fabrication, they will need to be used in conjunction with programmable hardware (most likely digital electronics, although it could also be analog) to allow them to be targeted towards specific tasks. If $>10^{12}$-parameter-scale foundation models could be executed cheaply using PFMs, they could be deployed to run locally in a variety of applications where this is currently not possible.}
    \label{fig:PFMs}
\end{figure}

\begin{figure}
    \centering
    \includegraphics[width=0.8\linewidth]{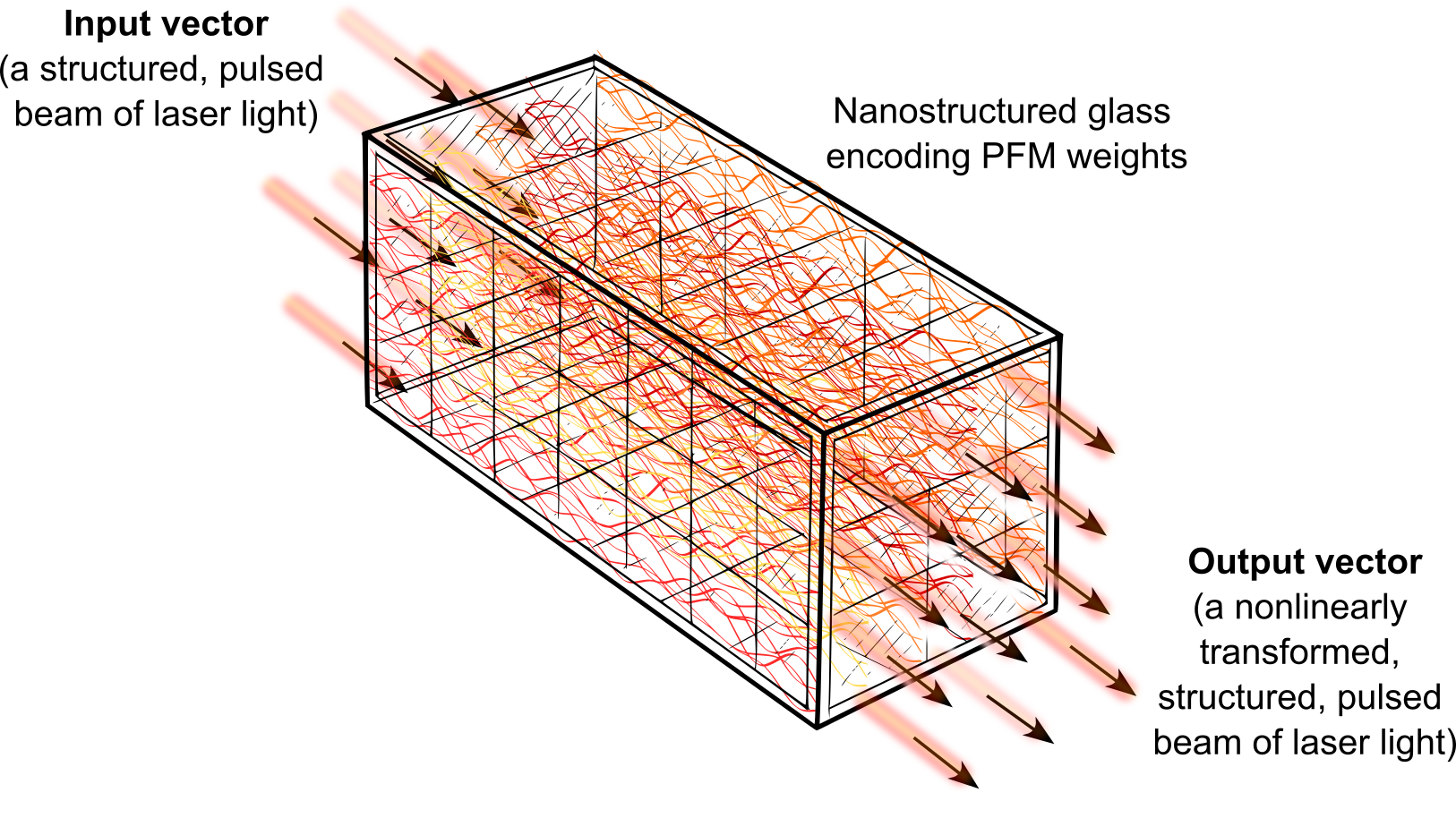}
    \caption{\textbf{Schematic of an example Optical Physical Foundation Model.} A 3D block of nanostructured glass encodes neural-network parameters in volumetric refractive index variations (with volume $\sim\lambda^3$ per parameter). Input vectors are encoded onto optical pulses via high-resolution spatial light modulators, and processing occurs by the propagation of these pulses through the medium. At high intensities, nonlinear optical propagation enables the deep, nonlinear transformations required for foundation models. The output is measured by a detector array and digitized. There are many other possible hardware realizations of PFMs, including in 2D and 3D analog electronics; more broadly, any physical system exhibiting complex dynamics that depend on many parameters that can be set at the time of fabrication is a candidate for constructing a PFM.}
    \label{fig:optical_pfm}
\end{figure}

\vspace{1ex}
\noindent \textbf{Why really big AI systems pose an unprecedented challenge for computer hardware:} We think there is an analog alternative worth exploring, but to see why let us first consider the scale of the challenge. Let us start with a $10^{18}$-parameter network. This is roughly a million times larger than today’s largest neural networks (${\sim}10^{12}$ parameters), which are in turn at least a million times bigger than late-1990s neural networks \cite{lecun1998gradient}.

A key observation is that $10^{18}$ parameters require a lot of memory to store---1 exabyte (assuming 8-bit parameters)---and a lot of memory bandwidth to access. There is no small system that can store this amount of data. At a storage density of 30~Gb/mm$^2$ (roughly the current state-of-the-art for 3D-NAND flash memory \cite{jung2024nand}) this number of parameters would require a physical memory area of ${\sim}$250~m$^2$. Even if we make the generous assumption that the scaling of layers of 3D flash memory can continue arbitrarily, so that memory area could be vastly reduced, the total volume would still be substantial\footnote{A back-of-the-envelope calculation assuming that a memory with 30~Gb/mm$^2$ density has a thickness of 50~$\mu \textrm{m}$ and is stacked $N$ times will have volume $V = (N \times 50~\mu \textrm{m}) \times (250~\textrm{m}^2)/N$.}: on the order of $10^4$~cm$^3$, corresponding to a cube larger than $20~\textrm{cm} \times 20~\textrm{cm} \times 20~\textrm{cm}$. This is far from cellphone-sized, and is large even for many robots.\footnote{It is worth emphasizing that an exabyte memory packed into a $20~\textrm{cm} \times 20~\textrm{cm} \times 20~\textrm{cm}$ volume does not currently exist, but is an optimistic projection of what might be possible with extensions to current storage technology. An exabyte storage system made from current solid-state drives, for example, would cost many million dollars and take up many m$^3$ of space.} Even if vast improvements are made in volumetric storage density, transferring such a large amount of data would be prohibitively time- (and energy- \cite{horowitz2014computings}) consuming. With a 1~Tbit/s memory bandwidth, it would take 90 days to read in the full model. It seems reasonable to conclude that neither inference nor training is likely to be practical with $10^{18}$-parameter models running on computers where the model memory is separated from the processing.

\noindent \textbf{Why might we even want to make hardware for AI models with $10^{15}$ or $10^{18}$ parameters?}

Do we really need models with $10^{12}$ parameters, let alone $10^{15}$ or $10^{18}$ parameters? Over the past few years, the approach of improving AI by (in large part) scaling the number of model parameters has---in spite of its amazing successes so far---been critiqued by various researchers (Refs.~\cite{levy2023arguments,marcus2025breaking} are two representative examples of critiques). We don't claim to know the answer to whether future AI will require using such large models. However, we would like to briefly motivate why we think developing technology that would enable running models with $\gg 10^{12}$ parameters could help advance AI.

\textit{Build it and they will come.} First, if training and running such large models were practical, people would probably do so. The trend in AI over the past decade has been of ever-larger models being developed as soon as it has been technologically and economically feasible to train and run them. More broadly, in the history of computing technology, there is a strong trend of new or expanded hardware capabilities being used to their fullest once they are made available.\footnote{Contra the apocryphal 1981 quote that ``640K (of memory) ought to be enough for anybody'', for the past ${\sim}$45 years, we have almost never failed to find use for larger memories, provided they were cheap enough.} However, if inference hardware capable of running, for example, $10^{15}$-parameter models at reasonable cost were developed, it would still not be adopted if training models for that hardware were infeasible. We discuss the bottleneck that training may pose and some possible solutions in Sec.~\ref{sec:discussion}.

\textit{Biology.} Second, there is the biology counting argument that $\gg 10^{12}$ parameters may be necessary to achieve human-like intelligence: human brains have roughly $10^{14}$ synapses, and given the substantially more complex behavior of neurons\footnote{Not to mention the additional complexity provided by dendrites, glial cells, etc.} in the brain~\cite{beniaguev2021single} than artificial neurons in current foundation models, one can reasonably hypothesize that achieving human-level intelligence may require models with $10^{15}$, or perhaps even $10^{18}$ or $10^{21}$, parameters. Achieving human-like (or even superhuman) intelligence is the goal of several prominent AI companies. Given what we know about the human brain, do we really expect $10^{12}$ parameters to be enough?

\textit{Spinoff benefits for hardware running smaller models.} Third, attempts to make inference sufficiently cheap (in energy, time, and \$ cost) for models with $\gg 10^{12}$ parameters will likely often result in the same benefits for models with $\lesssim 10^{12}$ parameters. We briefly discussed the potential impact of making inference at this scale more efficient in the previous subsection.

\subsection{A new approach: Physical Foundation Models}

In-memory computing \cite{ielmini2018memory,verma2019memory,mutlu2020modern,buchel2025analog} has been proposed as a solution to the von Neumann bottleneck---the bottleneck to computing speed because of limited bandwidth between a memory and a processor---even for far smaller models and memories than a current frontier model with $10^{12}$ parameters or a hypothetical future model with $10^{18}$ parameters, as we have described in the previous subsection. If we want to push toward ultra-large-scale models, the motivation for in-memory computing becomes ever stronger. However, programmable in-memory-compute platforms, such as matrix-vector multipliers based on programmable resistive crossbar arrays, have proven challenging to scale, with the largest storing just ${\sim}10^6$ parameters in a single array, and $<10^8$ parameters in a single chip containing multiple arrays \cite{ambrogio2023analog}. In addition, reliable and accurate programmability has been a long-standing challenge, as has write endurance \cite{haensch2023compute}.

There is, however, an alternative. Capitalizing on the emergence of foundation models \cite{bommasani2021opportunities} that exhibit zero-shot transfer learning, we note that programmability for a large fraction of AI inference hardware is not strictly necessary. Rather than attempting to construct programmable hardware, what if we designed single-purpose processors to execute foundation-model inference\footnote{The same question has been asked in some of the literature on compute-in-read-only-memory processors~\cite{chen2022yoloc,sehgal2023compute,yin2025hybrid,zhang2025bitrom}.} at the absolute maximum efficiency, using analog physical systems? We call these single-purpose analog computers \textit{physical foundation models}. Intuitively if we jettison the need to have programmability\footnote{Or programmability of most of the model parameters---preserving the ability to update at least a small fraction of the parameters.}, we will have reduced system complexity with potential benefits in parameter storage density, as well as energy efficiency, speed, and scale.\footnote{We elaborate on this in Sec.~\ref{sec:eliminating}.} Furthermore, by designing the hardware to execute the foundational model at the level of the physical substrate, rather than the traditional approach of going through many layers of abstraction and forcing the hardware to execute mathematical models that are not the most natural for the hardware, we can potentially gain orders-of-magnitude improvements in energy efficiency, speed, and parameter density~\cite{mead1990neuromorphic,sarpeshkar1998analog,wright2022deep}, in some cases even allowing the Landauer limit to be circumvented~\cite{mcmahon2023physics}.

\begin{redbox}{Physical Foundation Models}
\noindent A \textbf{\textit{physical foundation model}} (PFM) is a physical device that is made to perform a single computation---the inference of a trained AI model. Both \textbf{the algorithm (neural-network architecture\footnote{This is a simplification to avoid being too pedantic or generic while we are introducing the concept of PFMs: instead of hard-wiring a full neural-network architecture, one may wish to construct fixed-parameter physical-neural-network modules that can be connected in a programmable way.}) and the model parameters are \textit{hard-wired} into the physical device}, ideally using the lowest level of hardware physics possible (for example, the geometry and material properties of nanostructured semiconductors, which determine how electrons flow through the device) to achieve a compute-in-memory processor that has maximum parameter density and/or speed and/or energy efficiency and/or scale.
\end{redbox}

Figure~\ref{fig:PFMs} provides an overview of the PFM approach: a trained model is fabricated in fixed hardware which can then be deployed in datacenters for cloud access as well as in a variety of edge scenarios; to enable the foundation model to be used for a variety of tasks, it may need to be accompanied by task-specific frontend and/or backend processing, which could be performed by a set of hard-wired frontend and backend processors but in most scenarios will likely be programmable~\cite{yin2025hybrid}. This programmable processing could also potentially be used to compensate for discrepancies between the trained foundation model and the model realized in the physical hardware. In this paper, we sketch how PFMs could be made in optical and in nanoelectronic systems, and present back-of-the-envelope calculations for what speed and energy advantages these PFMs could have. We also discuss the open research challenges that need to be solved for PFMs to become a practical reality.

The research agenda that we describe here can be thought of, in part, as a merger of Feynman's in \emph{There's plenty of room at the bottom}~\cite{feynman1960there} and Leiserson~et~al.'s in \emph{There's plenty of room at the top}~\cite{leiserson2020there}: we are advocating for using the smallest possible feature sizes in device fabrication to directly literally write the parameters of the neural network into the physical hardware (Feynman), while getting rid of as much of the bloat in the architecture of modern digital electronic processors as possible by eliminating layers of abstraction (Leiserson et al.) and having the computation be designed directly at the level of device physics and dynamics. Intrinsic hardware-software co-design, i.e., a combination of hardware engineering pushed to the nanoscale and neural networks optimized for running on it, is not entirely new for AI \cite{brooksintelligence,hooker2021hardware,aifer2025solving}. Many critical recent landmarks in the field (such as AlexNet and Transformers) succeeded because they enabled unprecedented scaling of neural networks, in substantial part because they took advantage of the match between the calculations neural networks require and the calculations that can be most favorably scaled with available hardware (i.e., dense matrix-matrix multiplications and graphics processing units (GPUs)). The PFM approach is, in addition to using hard-wired rather than programmable parameters, to extend this hardware-software co-design---better matching hardware to neural-network calculations and vice versa---both to lower levels of hardware abstraction as well as to unprecedented levels of hardware specialization.

\section{How does eliminating (most) programmability lead to parameter-density, speed, energy, and scale advantages?}
\label{sec:eliminating}

PFMs are physical neural networks in which most, if not all, of the parameters are hard-wired rather than programmable. The potential benefits of implementing artificial neural networks at the level of physics has been covered extensively in the literature (e.g., Refs.~\cite{markovic2020physics,wright2022deep,yan2024emerging,momeni2025training}). Let us now briefly motivate the hard-wired aspect of PFMs.

\noindent\textbf{System complexity and parameter density.} Supporting programmability for any computing system, including physical neural networks, is generally not without cost: the system must be made more complex to allow programming, for example by the addition of data paths to transport programmable parameters to writeable memories; the addition of control paths to switch between writing and reading; and sometimes (as in the case of many non-volatile-memory technologies~\cite{yu2020compute}) additional complexity from the different operating conditions required to write versus read (e.g., write voltages being substantially higher than read voltages, necessitating additional circuitry to be able to generate and switch between higher and lower voltage ranges; writes in some memories needing complex feedback to correct for errors in the write process, whereas reads can be performed without any feedback). This additional system complexity usually directly results in an increase in system volume or area (versus an implementation that doesn't support programming), hence a reduction in parameter density. Ref.~\cite{yu2025dsc} summarizes an example of this in the context of compute-in-memory arrays, with programmable bitcells being $>10\times$ larger than read-only bitcells---and there are further density benefits from not needing the peripheral circuitry for writing. Especially for models with $\gg 10^{12}$ parameters, parameter density is an important metric, e.g., to enable the creation of physical neural networks that are physically small enough to fit in edge devices.

\noindent\textbf{Speed and energy.} Increased system complexity and hence increased system volume doesn't only affect parameter density. If the physical spacing between parameters is larger, the speed that the system can be operated at during inference time will generally be lower (e.g., from longer speed-of-light delay, or, in electrical circuits, from longer $RC$ delay because both parasitic resistance $R$ and parasitic capacitance $C$ increase with length), and the energy cost will generally be higher (e.g., in electrical circuits, due to more power dissipated in the higher parasitic resistance, and higher energy required to charge lines due to the higher parasitic capacitance). Another example from electronics of a possible energy benefit of storing fixed rather than programmable parameters is in the idling power consumption of read-only memory versus writeable memory, which often---as in the case of static random-access memory---involves more transistors (and hence more leakage current) or requires refreshing (such as in dynamic random-access memory).

\noindent\textbf{Scale.} Increased system complexity from programmability also often limits how many parameters the neural-network hardware can support in practice, for example due to the availability of sufficiently high-bandwidth interfaces to communicate the parameters to write, or due to constraints on maximum die size (increased parameter density then leads naturally to an increase in practically achievable scale). While limits to scale from system complexity are typically more about economics and what can be practically engineered given existing fabrication tools, packaging solutions, etc., as opposed to the limits to speed and energy, which tend to have more fundamental origins in physics, it is nevertheless plausible that going from programmable to fixed parameters could allow greater scale to be achieved in practice.

As a concrete example from optics, consider a metasurface whose pixels (which encode parameters) are fixed at the time of fabrication. A metasurface made from a 300-millimeter wafer can encode $>10^{12}$ parameters, assuming pixel dimensions of $\SI{250}{\nano\meter} \times \SI{250}{\nano\meter}$. In comparison, the commercial spatial light modulators with the most number of pixels have only $\approx 10^7$ pixels. Even if one restricts the comparison to devices of the same area, fixed metasurfaces can be realized with at least $100\times$ the number of parameters (i.e., $100\times$ greater scale) than currently available programmable spatial light modulators.

Enabling increased scale has the benefit of allowing larger neural networks to be implemented, which is desirable because it supports more intelligent behavior. However, increased scale can also enable increased energy efficiency: for example, analog matrix-vector multipliers in both electronics and photonics can feature an $O(N)$ scaling advantage in energy, where $N$ is the vector dimension~\cite{agarwal2016energy,shen2017deep,anderson2024optical}. An alternative motivation for why increased scale improves energy efficiency is that the energy cost of modern AI is typically dominated by data movement~\cite{verhelst2025keep}; if your AI processor can store the entire AI model's parameters in on-chip memory, it will generally be much more energy-efficient than a processor needing to perform many accesses of off-chip memory~\cite{hennessy2025computer}. The second-order effects on energy efficiency (due to increased scale) from abandoning programmability could well be even more important than the first-order effects.

\noindent\textbf{Parameter storage that doesn't exist in programmable form.} While it can be helpful for reasoning about the benefits of using hard-wired versus programmable parameters to compare using writeable memories to using read-only versions of these memories (as we have done in parts of this section so far), jettisoning the need to support the programming of parameters opens up the possibility of using device geometry or other physical materials or properties that usually don't get harnessed in memories (e.g., because they aren't amenable to electrical control) as a way to store neural-network parameters. Any aspect of a device's design that changes the device's behavior in responding to an input can be thought of as one or more fixed parameters. Directly using device geometry or material properties that can be set at the time of fabrication could lead to further advantages in parameter density, speed, energy, and scale---such as has previously been argued for fixed optical neural networks~\cite{khoram2019nanophotonic,onodera2024scaling}, but could similarly apply in electronics and other settings.

\section{Examples of possible Physical Foundation Models}
\label{sec:examples}

To assess the plausibility of PFMs, we performed back-of-the-envelope calculations for \hlnew{a nanophotonic 3D medium as a concrete example}. Appendix~A outlines detailed calculations and discussion of platform-specific issues, in each case considering a concrete model scale. These estimates are intended to be pedagogical, prioritizing transparency over design optimization. Our aim with these calculations is not to confidently predict the advantages achievable with PFMs, nor even to commit to a specific hardware realization. Rather, our aim is to verify that the premise for PFMs laid out here is physically realistic, at least at the level of back-of-the-envelope physics calculations. Our calculations, summarized in Table~\ref{tab:pfm_scaling}, imply that PFMs at the scale of $P=10^{12}$ to $P=10^{18}$ parameters are possible in principle, and offer the potential for many orders of magnitude improvement over current digital-electronic implementations for compact, energy-efficient, high-throughput inference. Although PFMs of increasing scale will clearly require increasingly ambitious fabrication, training, and simulation techniques, a key finding of our calculations is that the benefit of the PFM approach grows with model size $P$. Nonetheless, even at present-day scales of $P\approx 10^{12}$, PFMs appear to offer substantial advantages relative to programmable digital electronic implementations.

\hlnew{While we focus here on optics as a concrete example, nothing about the PFM concept is specific to optics. Electronic, mechanical, fluidic, or even chemical physical foundation models are all imaginable, though we expect optics, electronics, and mechanics are somewhat more likely to be scalable to the parameter counts required for foundation models. We discuss prospects for PFMs in other physical platforms, particularly nanoelectronics, in Sec.~\ref{sec:beyond_optics}.}

\hlred{\subsection{An example PFM based on optical wave propagation}}
\label{sec:optical_example}

\begin{table}[h]
\centering
\caption{\hlnew{Comparison of an Example Optical Physical Foundation Model (PFM) vs.\ a Digital-Electronic Baseline.}}
\label{tab:pfm_scaling}
\renewcommand{\arraystretch}{1.3}
\begin{tabular}{l|c|c|c|c|c}
\hline \hline
\textbf{Platform} & \textbf{Metric} & \textbf{Scaling} & \textbf{$P=10^{12}$ Params} & \textbf{$P=10^{15}$ Params} & \textbf{$P=10^{18}$ Params} \\ \hline
\multirow{3}{*}{\textbf{Optical PFM}}
 & Size (Volume) & $\sim P$ & $\sim (1\text{ mm})^2 \times 1 \text{ m}$ & $\sim (1\text{ cm})^2 \times 10 \text{ m}$ & $\sim (10\text{ cm})^2 \times 100 \text{ m}$ \\
 & Energy / Inference & $\sim P^{1/3}$ & $\sim 100~ \mu\text{J}$ & $\sim 1~ \text{mJ}$ & $\sim 10~\text{mJ}$ \\
 & Time / Inference & $\sim P^{1/3}$ & $\sim 100 \text{ ns}$ & $\sim 1~ \mu\text{s}$ & $\sim 10~ \mu\text{s}$ \\ \hline
\multirow{3}{*}{\textbf{Digital Electronics (Ref.)}}
 & Size (Volume) & $\sim P$ & $\sim (17 \text{ cm})^3$ & $\sim (1.7 \text{ m})^3$ & $\sim (17 \text{ m})^3$ \\
 & Energy / Inference & $\sim P$ & $\sim 1 \text{ J}$ & $\sim 1 \text{ kJ}$ & $\sim 1 \text{ MJ}$ \\
 & Time / Inference & $\sim 1$ & $\sim 25 \text{ ms}$ & $\sim 25 \text{ ms}$ & $\sim 25 \text{ ms}$ \\ \hline \hline
\end{tabular}
\end{table}

\noindent\textbf{General assumptions.} Our calculations make the assumption that a PFM is used within a broader digital-electronic system.\footnote{Users of ChatGPT, Gemini, Claude, etc. are still going to enter their prompts on digital-electronic hardware like phones or laptops, and are still going to have the results be returned to be displayed on the same hardware, even if the PFM responsible for much of the processing in the middle is neither digital nor electronic.} As such, the PFM has digital-electronic inputs and digital-electronic outputs, and is controlled by digital-electronic hardware. An important and yet dangerous simplification made in our calculations is that we only coarsely model many aspects of the digital-electronic system around the PFM hardware. These are physics calculations, not engineering designs. Most crucially we have estimated the costs of the systems around the PFM (e.g., cooling; digital-to-analog and analog-to-digital conversions; transport of input and output vectors) without considering or providing detailed designs for any of them. We have tried to make conservative estimates for these lumped costs (e.g., the cost of reading a vector element from digital-electronic memory, then converting it into the analog optical domain), but there are in many cases no existing hardware demonstrations of the peripheral digital and analog hardware PFMs will need at the scale they will inevitably require, so our cost estimates for the peripheral hardware should be treated as rough guesses, probably not even accurate to within an order of magnitude. With all analog calculations, we have assumed \textit{fixed output precision}, \hlnew{from the entire PFM in the optics case}. This assumption allows low-SNR internal operations that collectively sum to a finite-precision output, a key assumption common in many analog AI hardware proposals \cite{hamerly2019large,anderson2024optical,agarwal2016energy}, and an important one here to obtain energy scaling slower the digital reference.

\noindent\textbf{Optical PFM.} For the Optical PFM, we assume a 3D block of nanostructured glass where neural-network parameters are encoded in volumetric refractive index variations (with volume $\sim\lambda^3$ per parameter), as depicted in Fig.~\ref{fig:optical_pfm}. \footnote{\hlred{We adopt a 3D (volumetric) architecture deliberately: as our scaling estimates below illustrate, $\lambda^3$-per-parameter volumetric encoding is essentially required to reach $\gtrsim 10^{12}$ parameters in optics within manageable footprints, since planar (2D) optical neural networks devote $\sim\lambda^2$ of area per degree of freedom and quickly become unwieldy. The case~\cite{mcmahon2023physics} for 3D and volumetric optical neural networks has been advanced for decades by many in the optical-computing community, including by Psaltis and co-workers in their proposals for holographic neural networks~\cite{psaltis1990holography}, by Ozcan and co-workers in stacked diffractive deep neural networks~\cite{lin2018all}, more recently quantitatively by Li and Monticone using Miller-style~\cite{miller2023optics} information-theoretic arguments and explicitly invoking 3D volumetric metamaterials~\cite{li2025spatial}, and by the Microsoft team behind their recent analog optical computer~\cite{kalinin2025analog}. Our optical PFM example can be viewed as a foundation-model-scale extrapolation of this thesis.}} Thus, the PFM volume scales as $V\sim P\times \lambda^3$. This is, however, a bound on the required volume of nanostructured glass only---of course the required input modulators, output detectors, and laser sources would add additional volume, very likely enough to make portable Optical PFMs impractical. We assume inputs are encoded in optical pulses via high-resolution spatial light modulators, and that processing occurs by the propagation of these pulses through the medium. In the linear case (i.e., with relatively low power laser pulses), this structure would produce a simple matrix-vector multiplication (MVM), making it simply a 3D or continuous generalization of hardware we and others have proposed \cite{onodera2024scaling,nikkhah2024inverse,labroille2014efficient,fontaine2017design,kupianskyi2023high,lin2018all} (particularly Ref.~\cite{nikkhah2024inverse} and works on diffractive neural networks \cite{lin2018all} and multi-plane light-converters \cite{labroille2014efficient,fontaine2017design,kupianskyi2023high}, which either focus on or typically consider \textit{fixed} hardware). This MVM limit is useful as a lower bound on device performance, but of course a single-layer perceptron is not likely to be optimal for foundation models. As a result, we assume that input optical beams consist of short pulses with high peak power, such that their propagation through the nanostructured medium is $\textit{nonlinear}$, as in Refs.~\cite{hughes2019wave,khoram2019nanophotonic,nakajima2021neural,teugin2021scalable}, allowing the optical propagation to realize the deep, nonlinear transformations required for plausible foundation models.

As is commonly found for analog optical matrix-vector multipliers with this geometry, we find that energy consumption of this optical PFM can be assumed to be dominated by the input analog light vector generation and the output light detection and digitization, rather than the passive internal propagation. As a result, the energy cost of each inference scales with the larger of the number of input or output vector dimensions, as in optical analog matrix-vector multipliers. However, in the case of nonlinear propagation, this situation is slightly more complicated. Just as we can vary the number of parameters within a deep neural network without changing its width by simply adding more layers, in the nonlinear optical PFM we are free to (within realistic physical bounds) change the input and output dimensions without changing the number of model parameters, provided the propagation distance is long enough. Additionally, just as deep neural networks may use hidden layer widths that are larger than the input and output vector dimensions, we could likewise input lower-dimensional optical vectors than are allowed by the nonlinear optical wave propagation (that is, we inject or measure optical vectors in fewer spatial modes than are physically supported in the PFM). To resolve this choice, we assumed here that optical PFMs are realized as long tubes, essentially nanostructured multimode fibers. Although we expect optical gain to be required to maintain signal power (and nonlinearity) through the PFM, this fiber-like geometry is well-suited to minimizing the costs of this gain: total internal reflection of the fiber would minimize scattering of light outside the tube, and the geometry would ensure minimal need for cooling. With this choice of tube geometry, we then assume that the number of input and output dimensions are equal, and scaled as $P^{1/3}$, producing $10^4$, $10^{5}$, $10^6$ dimensions for the $10^{12}$-parameter, $10^{15}$-parameter, and $10^{18}$-parameter models respectively. Energy costs thus scale as $P^{1/3}$. The values in Table~\ref{tab:pfm_scaling} are computed assuming a cost per input and output vector element of 100 pJ, a lumped cost which includes all I/O, optical modulation, and the energy in the laser pulse\footnote{While our conclusions are insensitive to multiplying this number by another several orders of magnitude, we note that it is still a conservative estimate for these lumped costs, between 1-2 orders of magnitude larger than the lumped costs estimates in previous projections of optical neural networks, such as Refs. \cite{anderson2024optical,hamerly2019large}.}. Optical nonlinearities requires significant pulse energies and we find, for these lumped cost assumptions and our assumption of a fused silica medium, that the optical pulse energy required to access appreciable nonlinearity is the limiting factor on energy consumption (see Appendix). Since optical nonlinearity scales with intensity (power per unit area), the required optical peak power scales with the cross-sectional area of the PFM tube, and thus does not disrupt our assumed $P^{1/3}$ scaling. However, to maintain a high peak power over increasingly long tubes could require optical amplification and thus energy costs that scale with the length of the tube. Here, we have assumed that loss within the tube is minimal (e.g., by using only small refractive index variations within each glass voxel, minimizing backward and out-of-tube scattering), but this assumption is likely to hold less reliably for longer tubes.

The Optical PFM dimensions are computed by distributing the volume $V=P\times(\lambda/n)^3$ into a rectangular tube geometry. We allocated cross-sectional area and length to ensure sufficiently many guided modes (i.e., $M\sim An^2/\lambda^2$ should be equal to or larger than the assumed input/output dimensions) and otherwise arranged the design to produce minimally extreme dimensions in either cross-section or length. Here, we assume $\lambda = 1550$ nm, and $n=1.5$ for fused silica glass.

Finally, we find that the time per inference is limited by the total optical power in the optical medium, rather than by the propagation delay for pulses to transit through the medium. Assuming a 1 kW limit for the optical average power inside the PFM produces the inference times shown in Table \ref{tab:pfm_scaling}.

\noindent\textbf{Digital Electronics Reference (Baseline).} To contextualize these numbers, we include a Digital Reference based on state-of-the-art GPUs (e.g., NVIDIA B200 class). While we have applied a similar back-of-the-envelope style to this comparison, it should nonetheless not be considered as a completely direct comparison: GPUs differ from the example PFMs we consider in at least three key ways: the GPUs provide full programmability of the parameters as opposed to having fixed parameters; the GPUs are digital as opposed to analog; and the GPUs do not perform computing-in-memory whereas the example PFMs do.\footnote{The GPUs also exist as commercial, deployed products, whereas the PFMs we consider are currently purely hypothetical. This makes the comparison in some respect one of apples versus oranges, comparing something that exists with things that do not even have detailed engineering designs on paper yet. Nevertheless, we think it is useful to at least contextualize the potential of PFMs relative to current state-of-the-art. It would also be useful to compare the example PFMs we consider here (and other plausible PFMs) against fixed-parameter \textit{digital}-electronic hardware---which we might call \textit{digital PFMs}---to determine how much benefit can be gained from fixing parameters (and harnessing natural dynamics, and performing compute-in-memory) versus from operating in the analog domain. We leave this as future work, some of which is related to analyses performed for compute-in-read-only-memory processors~\cite{chen2022yoloc,sehgal2023compute,yin2025hybrid,zhang2025bitrom}.}

For our digital reference values, we begin by assuming a generous fixed energy cost of $\sim10^{-12}$ J (1~pJ) per 8-bit operation, then assume that the number of operations required to execute a model inference is equal to the number of parameters. For computing inference speed, we assume the GPUs operate at a fixed peak speed governed by high-bandwidth memory access, and we assume perfect parallelism where latency is bottlenecked by reading model weights from memory once per inference, ${\sim}$25~ms for 192~GB at 8~TB/s. These are optimistic assumptions: we are essentially neglecting communication overheads and their (usually unfavorable) scaling, and the immense complexity of interconnecting millions of processors, assuming that the GPUs all collectively operate near their peak performance. For size comparisons, we assume a single GPU occupies a volume of 1000~cm$^3$. This is approximately the volume of the card alone, neglecting the server chassis and cooling infrastructure. To allow direct comparison, we express the total volume of the millions of GPUs required for large models in terms of the side length of a hypothetical, perfectly packed cube.

\hlred{\subsection{PFMs beyond optics}}
\label{sec:beyond_optics}

\hlred{Here, we considered an optics example to make the case that PFMs are possible and could dramatically outperform conventional implementations of foundation models for inference. However, beyond this example, nothing we have assumed in this article is specific to optics---PFMs could instead be realized using nanoelectronics, nanomechanics, or even more unconventional platforms like microfluidics.}

\hlred{In preparing this manuscript, we additionally considered nanoelectronic PFMs under a variety of assumptions. Both human and AI-agent calculations for this setting can be maneuvered to yield promising scaling behavior, often arriving at the same conclusion as the optics case: that energy costs are dominated by input/output and nonlinear activations. Here, we briefly comment on each approach and describe why we have opted not to include them as quantitative examples. We then comment on routes we see as promising for realizing scalable PFMs beyond optics.}

\noindent\textbf{Crossbar arrays.} Crossbar arrays are a straightforward way to implement electronic circuits that realize matrix--vector multiplications, and formed one line of investigation. Implementing crossbars with fixed junction resistors does appear to be a direction worth further investigation as a means of realizing large PFM inference devices using much of the infrastructure of existing CMOS manufacturing~\cite{liu2025hnlpu,graf1988vlsi}. However, in our calculations on large-scale, fixed crossbars, we found that the scaling behavior of the energy and time required for large fixed matrices depended strongly on our assumptions---regarding interconnect dissipation, the scale of various complex sources of capacitive coupling, and so on---with different sets of assumptions yielding scaling of energy per inference ranging from $\sim N$ to $\sim N^4$, and thus performance that depends heavily on assumptions. In our best estimation, most or all fixed crossbar arrays would ultimately be dominated by parasitic wire dissipation scaling as $\sim N^4$ for the very large-$N$ matrices that would be considered in PFMs. While we do not completely rule out the possibility of a crossbar-array route to nanoelectronic PFMs, we consider the achievable performance to be sufficiently nuanced and assumption-dependent that we have opted to not include crossbar-array-based PFMs in Table~\ref{tab:pfm_scaling}.

\noindent\hlred{\textbf{Inverse-designed resistive wires.} Another route to nanoelectronic PFMs is to adopt the inverse-design philosophy of our optical example: patterning a 2D or 3D conductor's nanoscale resistivity (for example, by depositing metal or insulator in voxels several times larger than the electronic mean-free path). In this case, an important conclusion is that energy scaling is likely linear in the vector size $N$, avoiding the $N^4$ scaling encountered for crossbars. However, this linear scaling comes at the cost of unknown expressive power. While Tellegen's theorem~\cite{Tellegen1952} implies that such a passive resistor network is constrained to realize non-negative symmetric matrices, this does not mean that an arbitrarily patterned 2D conductor can realize \emph{arbitrary} non-negative symmetric matrices---in fact, this seems unlikely. In 3D, there is more geometric freedom and arbitrary non-negative symmetric matrices are more plausible, but heat dissipation and manufacturing are correspondingly more serious concerns. Operating in a superconducting transport regime would alleviate heat dissipation, but pose further challenges for manufacturing, especially as 3D structure would still likely be required for sufficient expressive power. In the limit of coherent transport, such as in a 2D electron gas, one can imagine realizing a structure almost precisely analogous to the optical PFM, relying on fixed electric fields to create a nanoscale potential landscape analogous to the optical refractive index pattern and thus accessing arbitrary unitary matrices with essentially dissipationless transport. This is, however, an exotic cryogenic regime that is highly speculative---it is not one from which we can confidently derive device performance bounds, let alone advocate as a scalable technology for neural-network inference (even if cryogenic overhead were minimized by performing inference in space or on the Moon).}

\noindent\hlred{\textbf{Inverse-designed semiconductors.} Another path to nanoelectronic PFMs would be to inverse-design semiconductor electronics: e.g., creating a 2D or 3D nanostructured medium of arbitrarily $n$- and/or $p$-doped semiconductor. This is an exciting hardware platform, certainly powerful enough to realize a wide range of analog neural-network-like transformations, including with manufacturing based on 2D CMOS processes. Encouragingly, recent experimental work on disordered dopant networks in silicon by van der Wiel and co-workers has shown that nanostructured semiconductors can host nontrivial nonlinear analog computations: small dopant-atom networks have been configured (via post-fabrication electrostatic tuning) to perform Boolean logic, classification, and other useful tasks~\cite{chen2020classification,ruiz2020deep,boon2021dnpu}. While these demonstrations are at far smaller scales than would be needed for foundation-model inference and rely on tunable rather than fully-fixed parameters, they suggest the promise of nanostructured semiconductors as expressive, manufacturable analog substrates and provide an early empirical foothold for the inverse-designed semiconductor PFMs we have in mind. Even so, back-of-the-envelope calculations here remain grossly insufficient to confidently assess perforance at scale, and inverse design would be extremely challenging: the coupling between electromagnetic, semiconductor drift--diffusion, thermal, and possibly coherent transport physics means this would be a formidable multiphysics problem for both theory and inverse design. Nonetheless, we are optimistic about this path.}

\noindent\hlred{\textbf{What physics lends itself to PFMs?} Three key desirable features are: (1)~that the physics be controllable and realizable at very small scales, so as to allow very large models to be physically encoded; (2)~that the physics be controllable by scalable, mass-manufacturable hardware; and (3)~that the energy cost of dynamics (information propagation and computation) in the physical system be low.}

\hlred{A further highly desirable---though we suspect not strictly necessary---property is that the physical process performing the analog computation can be reliably abstracted away from other physical processes occupying the same medium. Such an abstraction makes inverse design dramatically easier because it avoids needing to perform a fully coupled multiphysics simulation\footnote{Or a neural-network proxy of such a simulation.} (or to bridge a large simulation--reality gap arising from one). However, this clean separation of computational physics from other dynamics is not strictly required: even messy multiphysics platforms could plausibly be made tractable by inverse-designing small modules at a scale where multiphysics simulation is feasible, and then gluing those modules together into a larger system. This trades some efficiency for a much-reduced simulation burden, but still permits scalable PFMs to be built from physical substrates that lack a clean separation of computation from their other dynamics.}

\hlred{Optical wave propagation differentiates itself from electronics in being relatively abstractable from other physical processes in the same medium. This makes our back-of-the-envelope calculations somewhat more defensible, though they notably still ignore the inevitable thermo-optical effects that would arise in high-throughput use of the proposed PFM. There is a reason electronics has so universally adopted the digital paradigm, which achieves abstraction robustly despite the multitude of physical messiness in digital circuits. However, while the separation of optical wave propagation from other phenomena allowed us to make back-of-the-envelope calculations, it also showcases how the micron-scale wavelength of light necessitates 3D, inevitably macroscopic PFM hardware. Given the many advantages and inertia of 2D nanoscale manufacturing, scalable PFMs will therefore likely require either a 3D manufacturing paradigm, or alternative physical bases that can scale appropriately in quasi-2D architectures.\par}

\vspace{4ex}
\section{The future of physical foundation models}
\label{sec:discussion}

We now discuss what research challenges might need to be tackled to realize efficient, large-scale PFMs, and what impact PFMs and the effort to build them might have on AI and on computer science and electrical engineering more broadly.

\hlnew{The optics example of a PFM we have given highlights that the central challenge to realizing PFMs is one of large-scale physical inverse design and fabrication: How can and should we design large-scale (many-parameter) AI models directly at the lowest level of a scalable hardware’s underlying physics (e.g., wave propagation in nanophotonics)?} Many questions and challenges lie within this broader question:

\begin{itemize}
    \item \textbf{Training scalability beyond $10^{12}-$parameters: PFM networks-of-networks.} We think that if PFMs were realized, their advantages over conventional hardware would motivate novel approaches to model scaling and training that would allow for the creation of $10^{15}$-parameter or even larger models through engineered systems of coupled PFMs. But how would this be done, given that PFMs only improve inference efficiency and training $10^{12}-$parameter models is already extremely expensive? While the scale of models that can be trained will grow with digital infrastructure capabilities, PFMs offer a complementary route, namely using PFMs as specialist individuals within a vast network-of-networks, similar to mixture-of-experts models \cite{jacobs_adaptive_1991,shazeer_outrageously_2017,lepikhin_gshard_2021,fedus_switch_2022} and recent mixture-of-agents architectures that explicitly coordinate many LLMs \cite{wang_mixture-of-agents_2024}. PFMs could make this approach even more scalable. Training models consisting of thousands of individual PFMs would in this context then be feasible: it would consist primarily of training the digital managerial models and/or PFM network-of-network architecture, rather than training new PFMs themselves\footnote{In contrast to conventional differentiable MoE layers, where both experts and gating networks are updated by backpropagation~\cite{jacobs_adaptive_1991,shazeer_outrageously_2017,fedus_switch_2022}, the PFMs here would be fixed during this training, and non-differentiable. Training would thus takes place entirely in the digital ``manager'' and routing policies (and possibly in digital pre-/post-processing around PFMs), which must learn to exploit the fixed, heterogeneous behaviors of individual PFMs.}.
 In such a hybrid digital-analog context, the intrinsic variability of PFMs could be an advantage. While training thousands of distinct PFMs would be costly, even as a one-time expense, individually manufactured PFMs will, without extensive process optimization, exhibit variation due to the variation of their internal physical parameters within manufacturing tolerances (or due to variations in local environment). Since each fabricated PFM would exhibit variations from other nominally identical PFMs, assembling a network of thousands of distinct, individual PFMs might not require training thousands of distinct PFMs. Instead, training a network-of-networks could identify suitable specialist roles and teams from an ensemble of intrinsically variable fabricated PFMs, similar to ideas like model bagging and boosting \cite{breiman_bagging_1996,freund_experiments_1996,dietterich_experimental_2000}, but with manufacturing variability of individual PFMs replacing the deliberately induced variations in bagging and boosting, even if the PFMs were trained on identical data. This perspective even suggests deliberately engineering the fabrication process to be more variable, at least along functionally useful subspaces of the PFM parameter space, so that the ensemble spans a rich set of behaviors that downstream training of PFM networks-of-networks can exploit. In short, just as massive human projects succeed by way of coordinated teams of individuals, coupled through scalable, engineered collaboration systems and managerial oversight, scaling AI models with PFMs could take on a similar approach, relying on the extremely low cost of PFM-scale operations while using digital ``manager'' models to orchestrate large networks of PFMs.

    \item \textbf{Fabrication scalability}: In what physical platform can you practically make a compute-in-memory device that contains $10^{12}$ (or $10^{15}$ or $10^{18}$ or even $10^{21}$) parameters? If you need to use 3 dimensions to make the device reasonably compact (which seems likely, at least for $>10^{15}$ parameters), or possibly more energy-efficient~\cite{boahen2022dendrocentric}, how will the fabrication be done at scale?
    \item \textbf{Physical computation constraints and expressivity}: What physical platform's underlying physical dynamics gives rise to computations that are most useful for performing the kinds of computations needed in foundation models \cite{wright2022deep}? What is the expressive power \cite{raghu2017expressive} of the physical dynamics, and how efficient (in space, time, and energy) is it at computing desired functions? How much should the physical dynamics of a platform be constrained so that the computations it performs better match computational primitives that are useful for foundation models?\footnote{\hlnew{An extreme example is imposing a structured architecture (such as a crossbar) to constrain the platform to performing matrix-vector multiplications.}}
    \item \textbf{Tolerating or harnessing probabilistic/stochastic dynamics}: While we have focused the examples and discussion in this article on analog computing \cite{huang2017hybrid,achour2020noise,wright2022deep,momeni2023backpropagation,wang2023design,song2024programming} that is usually performed with relatively high signal-to-noise ratios, PFMs could also be engineered with much noisier systems, including ones exhibiting probabilistic discrete behavior \cite{kaiser2021probabilistic}, which often emerges when operating at quantum limits of energy consumption \cite{ma2025quantum}. Physical stochastic hardware can implement approximate \cite{han2013approximate}, stochastic \cite{alaghi2013survey}, or probabilistic \cite{palem2012what} computing at the level of physics---for example, using devices whose relevant behavior is determined by statistical thermodynamics or quantum mechanics. For some algorithms and applications, stochastic operation is not merely something that can be tolerated but rather a core part of the algorithm \cite{mitzenmacher2017probability}---for example, in generative models using diffusion \cite{sohl2015deep,luo2022understanding}---that is natural to engineer directly at the physical layer~\cite{conte2019thermodynamic,coles2023thermodynamic}. How could stochastic hardware benefit PFMs through implementing algorithms where randomness is useful, or through trading off accuracy for energy or speed benefits? What computational expressivity is possible with various choices of stochastic hardware? How should stochastic and quasi-deterministic hardware be combined to realize PFMs that have stochastic operation in some parts but perform deterministic computation in others?

    \item \textbf{Neural-network architecture}: In analogy to how large-scale deep neural networks are given some structure (architecture) before training, which helps in allowing them to scale, PFMs may benefit from (or even require) some structure being imposed on their physical structure before training. However, this structure probably won't be the exact same one that is appropriate for neural networks implemented in digital-electronic processors using matrix-vector multiplication, and we need to discover what it is (for different choices of physical platform and dynamics). Which aspects of the architectures of current digital-processor-based foundation models (for example, attention in Transformers~\cite{vaswani2017attention}) should we try to take inspiration from, and which should we jettison?

    \item \textbf{Modularity and hierarchy}: Should we really try to train the full (extremely large) PFM all at once---treating it as a monolithic computation, or should we introduce some modularity to make both fabricability and training easier?\footnote{If modules could be trained separately, without needing information from the rest of the PFM during their training, the simulation burden could be dramatically reduced. Alternatively, if training could be performed in a way that builds up from training a single module alone, whose parameters are then fixed, to training the modules connected to the first module, and then the modules connected to those modules, and so on (Ref.~\cite{momeni2025training} surveys some training approaches like this in its section on \emph{physical local learning}), then one could fabricate physical implementations of the modules that have already been trained and have those physical implementations help to accelerate the training of the not-yet-trained modules. In both scenarios, one could avoid needing to run the full model in conventional hardware.} How and where should we introduce modules? Should there be a hierarchy of modules? What, if any, connection should there be between the network architecture and the modules?
    \item \textbf{Simulation infrastructure for training}: Training a PFM is essentially an exercise in large-scale inverse design with a machine-learning objective function. How can the training be performed in a way that doesn't cost too much time, energy, or money? Training of conventional foundation models already limits their size and how quickly new models can be released. For PFMs to be impactful, the additional difficulties one faces in training them must not outweigh the advantages they confer in inference. While some choices for PFM hardware, \hlnew{such as a physical system constrained to implement a matrix-vector multiplication, might perform computations that are relatively inexpensive}\footnote{Inexpensive per trainable parameter (e.g., one or two or three arithmetic operations per trainable parameter), not inexpensive in total---the topic of this paper is neural networks so large that that are expensive to perform inference with, so the full computation performed by the hardware must be expensive.} to simulate on a digital-electronic computer \hlnew{(such as matrix-vector multiplication)}, most physical systems will be expensive to directly simulate---so conventional approaches to gradient-based inverse design will be expensive \cite{lu2021physics}. How can simulations of physical platforms be made sufficiently inexpensive that they are practical for very large-scale ($\gg 10^{12}$-parameter) systems, but also sufficiently accurate that fabricated designs based on those simulations perform (approximately) the desired computations? How accurate is accurate enough? Fortunately the inverse-design community more broadly faces the same challenges from simulations being impractically expensive for large-scale designs, so techniques from them can be adapted---such as the use of surrogate neural-network models to replace direct simulations \cite{lu2021physics}.

    Beyond some number of trainable parameters (likely between $10^{12}$ and $10^{18}$), even highly efficient simulations running on digital-electronic computers will be too expensive to be practical.\footnote{Simply because one should need at least one arithmetic operation per trainable parameter to simulate a forward pass, and, for example, executing $10^{18}$ arithmetic operations is expensive in both time and energy.} How then will it be possible to train ultra-large-scale PFMs with, for example, $10^{15}$ or $10^{18}$ parameters?\footnote{This is a problem we would welcome having: already being able to build PFMs with $10^{12}$ or $10^{13}$ or $10^{14}$ parameters could be transformational, so even if that is as large as one can go with simulation-based training, it could already be a major success.} One possibility is that existing PFMs will be able to aid in the design of future PFMs, e.g., by being able to predict gradients, allowing the bootstrapping of ever-larger PFM designs.\footnote{This is in analogy to how current digital-electronic computers are used to design even more powerful future digital-electronic computers.} Another promising direction is to develop PFMs whose parameters in hardware are not programmable from outside the PFM but that are also not fixed at fabrication. Here the parameters could change according to internal dynamics within the hardware \cite{momeni2025training}, potentially taking advantage of novel materials and circuits with memory \cite{sangwan2020neuromorphic,markovic2020physics}. This direction is complementary to some of the ideas discussed below (under \textit{Fabrication variability, and simulation--reality gap}) for introducing post-fabrication tuning to compensate for differences between the behavior of fabricated devices and simulations of them.
    \item \textbf{Compilation}: Is there a way to transfer parameters from a conventional neural-network model (like GPT-5) directly to a hardware design? While not necessary for PFMs to succeed, if such a technique existed and was sufficiently inexpensive, it would dramatically lower the barrier for PFMs to be adopted.
    \item \textbf{Fabrication variability, and simulation--reality gap}: Training (or compiling) will likely involve making predictions about the behavior of a hardware design, typically through simulations or neural-network surrogates for simulations \cite{lu2021physics}, but those predictions will never perfectly match the behavior of fabricated hardware. What should be done to ensure that, in spite of the gap between simulation predictions (upon which the training is based) and fabricated-hardware reality, the PFM hardware achieves the correct neural-network behavior? For small deviations, many of the techniques developed for training and deploying neural networks to noisy analog hardware \cite{zhu2020statistical,semenova2022noise,hamerly2022asymptotically,vadlamani2023transferable,anderson2024optical,buchel2025analog} will likely be applicable. However, if the hardware is pushed to achieve impressive parameter density and computational expressivity at large scale, it is likely that accurate\footnote{Meaning giving a good---albeit not exact---quantitative match between simulation predictions and hardware behavior, e.g., to the level where the mismatch can be dealt with by treating the hardware's behavior as a slightly noisy version of the simulation's.} simulations will become impractical to make or run. Consequently realizing large-scale PFMs will likely need new ideas for how to deal with the large simulation--reality gap. One direction is to add small digital-electronic neural networks before or after the PFM hardware, which can be trained to compensate for the mismatch in behavior between the simulated and fabricated PFM. If each fabricated copy of the PFM device has the same behavior, then these compensation neural networks need only be trained once with one copy of the PFM, and all other copies can use the same compensation parameters. If there is copy-to-copy variability in the fabrication, then each PFM copy may need to have the compensating digital neural networks be trained separately. An alternative or complement to sandwiching a PFM inside digital neural networks to compensate for variations is to add some amount of post-fabricating tuning to the PFM hardware itself. For example, imagine that not all the parameters in the PFM are permanently fixed during fabrication, but that a small fraction are reprogrammable~\cite{chen2022yoloc,taniguchi2025alora,yin2025hybrid}.\footnote{\hlnew{For example, in a nanoelectronic architecture, some fraction of parameters could be realized with reprogrammable elements such as memristors.}} It is plausible that having only a relatively small amount of reprogrammable parameters could have a large overall effect, as has been demonstrated in the context of fine-tuning large models to perform specialized tasks~\cite{hu2021lora}. This post-fabrication tuning may also be possible to do with forward passes only through the PFM hardware itself, making it inexpensive relative to the original training \cite{malladi2023fine}. How best to introduce some reprogrammability into different physical platforms for PFMs is an open question. Needing such post-fabrication tuning isn't desirable but could still be acceptable if the advantages of PFMs are large enough. This has recently been argued by Hinton, who termed \textit{mortal computation} as the situation in which each copy of neural-network hardware needs its own unique parameters\footnote{Which are discarded when the hardware dies, hence \textit{mortal}.} \cite{hinton2022forward}: ``\textit{If you want your trillion parameter neural net to only consume a few watts, mortal computation may be the only option. Its feasibility rests on finding a learning procedure that can run efficiently in hardware whose precise details are unknown}''. How to efficiently perform the post-fabrication tuning (training) is also an open question, and can draw approaches from the community working on methods to train physical neural networks more broadly \cite{momeni2025training}.

    \item \textbf{Robustness to---and/or mitigation of---fabrication defects, especially catastrophic defects}: \hlnew{When fabricating PFMs with $>10^9$ internal parameters, many designs will suffer} from the possibility of \hlnew{a fabrication defect causing catastrophic consequences (e.g., a short circuit in electronic implementations, or a scattering defect in optical ones)}. Large memories (such as dynamic random access memories) in conventional computers face similar challenges and require mitigation of fabrication defects, often by providing some redundant capacity and using post-fabrication modification with fuses~\cite{horiguchi2011nanoscale}. How can PFMs (nanoelectronic or otherwise\footnote{The short-circuit example given here doesn't apply to all possible physical platforms, but many physical platforms will have the potential for fabrication defects to cause some catastrophic consequence in certain designs.}) be engineered to be robust to fabrication defects? Can PFMs be designed to be intrinsically robust to defects without losing their advantages? Or will they need post-fabrication testing and modification? If so, how can this be done without being excessively costly or slow?

    \item \textbf{Disposability and repair of PFMs}: The production and deployment cycle we suggest for PFMs could lead to the accumulation of trash devices, as each new generation of fixed device is deployed, and because PFMs may be more fragile and difficult to repair than digital devices. How could this trash accumulation be mitigated, and are there realistic paths to repairable PFMs? While the same concerns apply to present-day computer hardware (and may form a necessary part of the associated business model), PFMs could both extend or diminish this challenge. For example, PFMs like the examples we have considered in our calculations are constructed of very few materials---nanostructured glass or metal, for example. If practical PFMs retain this material simplicity, recycling them could be similarly simpler than other hardware: they could merely be melted down. On the other hand, the analog degradation of PFMs could make their safe operating lifetimes shorter than digital devices, so rapid and efficient turnover may be a necessity rather than a bonus feature. Repair and reverse engineering of PFMs would be practically impossible at microscopic scales (even more difficult than current hardware). However, if they are designed and constructed in a modular fashion, errors in individual modules could be diagnosed, and the corresponding modules could be replaced. As mentioned already, such modular design would also make manufacturing PFMs and compiling them from existing models less expensive and more interpretable, so modular repairs (and modular hardware modifications) are likely to be possible. Further, because they are essentially embodied realizations of neural networks, PFMs would likely be amenable to forms of repair that qualitatively differ from current hardware, and that could be realized through external means, such as modified prompts or retraining of pre- and post-processing layers.
    \item \textbf{Parameter precision--volume tradeoff}: There will be a tradeoff between how small we can make the volume storing a single parameter and what precision we can have for the parameter encoded in that volume.\footnote{This might be too simplistic of a picture, although it gets at the spirit of the tradeoff: in many physical systems, one can imagine that shrinking the volume used to store parameters may cause crosstalk or other deleterious effects that manifest as, for example, undesired correlations between parameters, or undesired physical dynamics---effects that (in some cases) may go beyond a mere reduction of effective precision.} What is the optimal choice when considering system-level metrics such as inference accuracy, speed, and energy?

    \item \textbf{Interfacing}: How should we design PFMs to interface with traditional digital electronics efficiently?\footnote{For example, if digital-to-analog and analog-to-digital converters are used, one wouldn't want to need so many of them that their energy costs become prohibitive.} How should PFMs interface directly with analog signals when used as part of smart sensors \cite{ballard2021machine}?
\end{itemize}

\subsection{Technological and scientific opportunities enabled by PFMs}

If the research vision that this paper outlines is successful, we would have cheap, low-power hardware capable of executing large foundation models. This could have important consequences beyond ``just'' reducing the power consumption of current foundation-model-based cloud-computing services (such as ChatGPT):

\begin{itemize}
    \item \textbf{New computer architectures and software stacks}: If computers can each have a cheap, low-power foundation-model co-processor, we can envision a shift in conventional digital-electronic computer architectures (and the software stack built on them). With essentially free access to a foundation model, we can imagine new computers being designed that accelerate the adoption of neural-network-centric software \cite{karpathy2017software,welsh2022end}.
    \item \textbf{New edge capabilities}: If foundation-model processing is cheap, it could be used in many different edge devices that either don't have network connectivity or that require low latency, potentially enabling a whole new class of edge-device capabilities. There is also a natural interpolation between now and when (if) low-power, large-scale PFMs become available: progress toward large-scale physical neural networks that can execute foundation models will naturally give rise to small- and medium-scale physical neural networks that aren't big enough to run large foundation models (like GPT-5 or Gemini~3 or Llama~4) but can run task-specific neural networks at low power \cite{xu2018scaling}, or smaller-scale foundation models tailored to edge tasks such as sensor processing \cite{abbaspourazad2024largescale,narayanswamy2024scaling}.
    \item \textbf{Improved security}: In addition to potentially enabling foundation-model computation without needing to use an external cloud service, supporting data privacy, there is also the unusual feature that (most of) the foundation model is fixed in hardware at the time of fabrication, which reduces the potential for malicious modification of the neural network.
    \item \textbf{An opening to use unconventional hardware more broadly}: There is currently well-justified reluctance to adopting unconventional hardware to augment CMOS-based digital-electronic computers.\footnote{One common and reasonable argument against making the effort to integrate unconventional coprocessors into computing systems is that there have been multiple waves of interest in both electronic and optical analog computers over the past several decades and none have resulted in widespread commercial impact. We believe this time could be different, but we understand the skepticism.} If PFMs are introduced and adopted, this may encourage the adoption of unconventional computing hardware in other parts of computing systems.
    \item \textbf{Application of design techniques beyond computing}: The techniques developed to enable the design of PFMs could potentially enable large-scale inverse design of physical hardware at a low level for functionality besides computation---such as communications, sensing, and actuation---and systems that might combine all of these, such as nanorobots.
\end{itemize}

\subsection{Capabilities and control of PFM-based AI systems}

PFMs would bring both new opportunities and risks beyond those already presented by modern foundation models \cite{bommasani2021opportunities}. They would allow for radical scaling of AI systems, and would bring these systems both into new, analog physical substrates as well as new environments. We do not know what capabilities AI systems of this kind and scale will have, nor what their vulnerabilities may be. The research direction PFMs represent thus poses challenges for AI safety, and will require more research focused not just on technology development, but also understanding. To this end, although PFMs pose many qualitatively new challenges for understanding and safely, productively applying AI systems, they also appear to offer qualitatively new opportunities for researchers and engineers to understand and control these systems. In this section we outline why we believe PFMs deserve the attention of experts on these topics, and the questions we feel need to be addressed. We are applied physicists, far from experts in the rapidly developing fields of AI understanding and safety, so our short answer to all these questions is simple---we do not know. Nonetheless, we hope here to provide some clues (and motivation) for the necessary research that will be required to address them.

\begin{itemize}

\item \textbf{What makes PFMs different from conventional digital AI systems?} First, PFM-based AI systems would be primarily fixed, with a seemingly limited capacity for reprogrammability or learning. Second, they would be (mostly) analog: their computations would be performed (and designed or compiled) at the level of microscopic, analog physical processes. PFM-based AIs would be grounded in physical processes, and individual and mortal \cite{hinton2022forward,hinton2024romanes}, to use Hinton's term. Third, they would allow AI systems of vastly larger scale and greater efficiency. Finally, they would allow these AI systems to be deployed not only in remote-accessible hardware, but locally and privately in a variety of edge devices.

\item \textbf{How powerful would PFM-based AI systems be?} Our back-of-the-envelope calculations suggest that PFM-based AI systems could reach the scale of $\sim10^{12}$ to $\sim10^{18}$parameters, plausibly with energy consumption amenable to portable, edge implementations. These models could thus be larger than the largest present-day models (known or estimated to have $\sim10^{13}$ parameters) by $10^2$ to $10^5$ times, comparable to the ratio between GPT-2 and GPT-4 ($\approx10^3$), and Bengio~et~al.'s neural probabilistic language model \cite{bengio2003neural} and GPT-4 ($\approx10^6$) respectively. Moreover, PFM-based AI systems would be able to perform inference more efficiently than present-day systems---\hlredbf{here our calculations (Table~\ref{tab:pfm_scaling}) suggest energy per inference between $10^4$ and $10^8$ times smaller than current digital hardware at matched parameter counts (depending on size), so that even a $10^{18}$-parameter PFM would consume roughly $100\times$ less energy per inference than today's $\sim 10^{12}$-parameter digital systems}. As a result, PFMs could, possibly even at smaller scales, allow also unprecedented \textit{test-time scaling} \cite{cobbe2021training,epoch2023tradingoffcomputeintrainingandinference,snell2024scaling,balachandran2025inference,wu2024inference}. This is particularly true if digital-analog conversions could be avoided between inferences, as PFM-based AI's power consumption will likely be limited by this step. Although it is in part misleading to directly compare these analog operations and parameters with the digital ones of present-day AI systems, it seems very likely that PFM-based AIs would be much more powerful than current systems.\footnote{Since $10^{12}$-parameter (and larger) PFMs don't exist yet, we also need to consider how they would fare against future conventional systems in 10 or 20 years. Setting aside the contribution of AI model (software) advances---which might well also be applicable to PFMs---to the capabilities of future AI, and focusing only on hardware capabilities, PFMs should be able to achieve a large advantage even against GPUs 20 years in the future, under the assumption that GPUs will improve in energy efficiency and performance by less than $1.5\times$ per year~\cite{epoch2024priceperformancehardware}.}

\item \textbf{What capabilities would PFM-based AI systems have?} Neural scaling laws \cite{Hestness2017,kaplan2020scaling,Bahri2024,Hoffmann2022} imply an aspect of predictability to the performance of large AI models, but we have far less understanding of the qualitative and/or emergent capabilities such scaling could enable. It may be that such models would be sufficient to perform vastly superhuman feats of seemingly unlimited domain, but it is also possible that they will only be sufficient to allow AIs of the kind we currently access to be widely, reliably, locally, and profitably deployed. In edge settings such as within robots or cars, these AI systems will need to navigate a world that is, at least intuitively, much higher dimensional than language, and so AIs that perform reliably in these settings may need to be significantly larger scale than those required to achieve similar (e.g., human-relative) performance with text. Thus, PFM-based AI systems might still be insufficient to allow intellectually flexible robots that can effectively assist humans at subtly complex tasks like caregiving or driving. To provide some context, it is tempting to compare AI systems to a familiar ``capability reference point''. Human brains, abstracted as neural networks, contain about $10^{14}$ neural synapses, suggesting that perhaps PFM-based AIs could be quite human-like in their capabilities. It may, however, be more appropriate to think of these $10^{14}$ synapses as merely representing the reprogrammable parameters of our brains, a small subset of its evolutionarily optimized parameters. We can thus repeat the traditional mistake of equating the brain to the latest technology paradigm, now to a physical foundation model, and note that our brains have evolved as spectacularly complicated, optimized analog machines, containing $\sim10^{12}$ diverse individual cells, with each cell a bewilderingly complex, optimized analog machine in its own right, containing within itself $\sim10^{12}$ coordinated biological nanomachines (proteins). But even if $10^{24}$ parameters (i.e., counting the number of coordinated proteins in the brain) is the proper reference for human-brain-scale AI, PFM-based AIs could operate at \hlredbf{significantly higher speeds than our brains (perhaps $10^3$ times, as suggested by our calculations summarized in Table~\ref{tab:pfm_scaling})}, and could be trained to specialize in a single class of tasks (like today's mostly language-oriented models), so they might outperform humans in many tasks. In short, while the incredible scaling potential of PFM-based AIs seems to at first imply equally incredible capabilities (and we think this possibility should be considered seriously), we do not think it is obvious that PFM-based AIs would dramatically outperform humans, particularly in the physical world or mixed-domain settings. In short, our uncertainty is very high: These are important questions for further research.

\item \textbf{What new challenges would PFM-based AI systems present for the understanding and safe management of AI?} In addition to enabling potentially new capabilities in familiar settings for AI, PFMs could enable widespread deployment of AI systems in a wide range of contexts, and unforeseen dangerous behaviors could arise in these settings. In these settings, errors which are minor nuisances for users of language models today could be devastating. Further, PFM-based AIs would be \textit{individual}: manufacturing variations, or damage to their internal components over time, would mean that each system would be slightly different, as we have assumed their calculations are performed at the lowest possible physical scale of their hardware. Damage, such as from cosmic radiation or cycles of external temperature variation, would make PFM-based AIs \textit{mortal} \cite{hinton2022forward} and would likely pose significant challenges for current methods of controlling AI behaviors such as human-feedback reinforcement learning \cite{christiano2017deep,bai2022training} and machine unlearning \cite{barez2025open}, necessitating careful tracking of their performance (and eventual repair or replacement). Measures to manage individuality and mortality of critical technology products, such as regular testing of cars, could be adapted in part to PFM-based AIs, but only incompletely because we currently do not understand what important failure modes (and their diagnostic symptoms) would be. The limited programmability of PFM-based AIs could also make it difficult to efficiently correct for such errors without physically modifying them. Overall, irrespective of whether PFM-based AI systems would enable vastly superhuman AI, the analog fragility and variability of PFM-based AI systems would make them unpredictable and potentially dangerous without careful management.

\item \textbf{What opportunities do PFM-based AI systems present for the understanding and safe management of AI?} PFM-based AIs would be mostly fixed, and thus largely incapable of learning during their lifetimes. They would be individual, and thus would not be capable of the same dramatic parallel learning of cloud-accessed digital AIs \cite{hinton2024romanes}. These mean that the evolution of PFM-based AI systems might be easier to regulate, since learning would primarily be a generational process more analogous to the design of new hardware, which could be overseen by humans, rather than a distributed and continuous process. To partially (and possibly, more controllably) recover this parallel-experience learning, these analog AIs could be designed to keep track of key surprises or ``lessons'' throughout their operational lifetimes, providing efficient training data or harnesses for the next generation. To regulate learning, this data could be curated prior to model updates, and the resulting AIs could be systematically tested \cite{grey2025safety} before releasing them in the form of PFMs. PFM-based AIs would also be physically grounded. Software-based AI systems rely on mathematical operations that can be changed (e.g., by a self-modifying AI coding system), and as a result, conclusions from analysis of their operation at this level of abstraction (and many imposed constraints) could also be changed. In contrast, hardware-based physical AI systems like PFMs are implemented in the physical evolution of their analog substrates, giving hope to the possibility that their operation might be understood, designed, and constrained by inalienable physical laws. The partially blind development of modern AI systems has been likened to the early era of steam-powered machinery, advancing intuitively and empirically, but still poorly understood \cite{venkatasubramanian2025large}. Physically grounding AI systems, like steam-powered machines, could catalyze the discovery and application of physical theories of AI that include fundamental constraints, analogous to those universal fundamental limits on engine efficiency that were subsequently derived using the then-new field of thermodynamics. It is this combination of constraints---those of the fixed hardware and its essential physical embodiment---that to us (admittedly, physicists) inspires the most potential for advancing capable, reliable, safe and understandable large-scale AI systems.

\end{itemize}

\subsection{Why do PFMs make sense to study now?}

We see three reasons. Firstly, there is enormous and growing demand for efficient AI inference on ever-larger models. Secondly, whereas before the advent of foundation models in the past ${\sim}$4 years \cite{bommasani2021opportunities}, it was almost nonsensical to think of a neural-network processor whose model parameters couldn’t be changed, now it is reasonable to consider processors where the vast majority of parameters are fixed at fabrication time. Finally, it has become apparent that deep neural-networks don’t necessarily have to follow the conventional structure developed in the artificial-neural-network community, and that co-designing the physical substrate and the network may lead to expressive, accurate neural networks that are also extremely well-matched to hardware \cite{wright2022deep,jaeger2023toward,laydevant2024hardware,momeni2025training}.

\section{Summary and Conclusion}

We end with a summary, and a final justification for why we believe our speculations here are perhaps worth pursuing. In this article, we have sketched an approach to computing hardware that embraces the unusual nature of large AI foundation models---namely that they may be used in an extremely wide range of applications, but yet in each application they primarily perform the same calculations, just with different input data. This, in principle, allows us to specialize hardware to an extent that would in most other contexts seem absurd: to the single purpose of performing foundation-model inference, eschewing programmability entirely in exchange for energy efficiency, speed, and compactness vastly beyond what is possible in conventional programmable digital-electronic hardware. We call these single-purpose analog computers \textit{physical foundation models}. Scaling current state-of-the-art neural networks by several orders of magnitude will, most likely, require a radical change in how they are implemented in hardware---even at their current scale, the compute resources and energy required are pushing the limits of what is practically viable. PFMs provide a possible path to scaling to neural networks with $10^{15}$ or even $10^{18}$ parameters and beyond, without inference becoming prohibitively costly.

While this new hardware will inevitably need to apply many of the lessons we have learned from digital artificial neural networks, we think it is probably not necessary or optimal to strictly copy the precise form (architecture) of modern neural networks, such as Transformers. The reason is that the optimal architecture depends on its physical implementation and vice versa. Your brain is able to perform many of the same functions as modern neural networks, but it does so using an arrangement of microscopic analog machines whose precise mathematical description resembles modern artificial neural networks only in a very coarse-grained sense. There are, for example, no ReLU activation functions involved in you reading these words right now. Hence, while we could \hlnew{realize analog AI systems in faithful emulation of these architectures, carefully approximating each individual operation of a modern neural network with a corresponding physical system, this is unlikely to be the best way} to use the physical machinery we have to realize the global functions we wish to perform with AI systems. While Transformers have proven to be an exceptional match to the efficient operations of GPUs~\cite{hooker2021hardware}, it will likely be a different, more microscopic-physics-aware mathematical ansatz\footnote{A parameterized mathematical form chosen as a starting point for optimization or learning. In this context, an ansatz specifies the class of functions the model can represent---analogous to the role of architecture in conventional neural networks.} that is optimally scalable to realize the most compact, high-throughput and energetically efficient AI inference ~\cite{wright2022deep}---for example, \hlnew{by the propagation of light through nanostructured glass, or the flow of electrons through nanostructured electronic materials}. In other words, achieving powerful machine-learning capability while also seeking efficiency in the physical compute hardware is a joint optimization, a compromise between what is scalable and efficient in hardware, and what works at scale in software.

Consequently, the goal we are advocating for boils down to designing an analog computing machine architecture that is scalable in its complexity, reliably manufacturable (to the extent necessary), and whose microscopic parameters can be principally learned using gradient descent (or any other adequately scaling training algorithm). We could opt to fix one part of this optimization, at today's artificial neural network's mathematical forms, in Transformers and the like, and then design application-specific integrated circuits specifically for these forms. Or, we could simply let the optimization extend as far as it can into the hardware itself, optimizing not weights of an abstract mathematical model ans\"atze, but instead those of the physical system---the hardware itself. This joint optimization leads to the hardware and software (model architecture and parameters) being inseparable: as in biology, the hardware becomes the software~\cite{laydevant2024hardware}, and vice versa. Put another way, PFMs are an example of \textit{physical computer science}: computer science practiced at the level of physical hardware \cite{jaeger2023toward,valiant2024matrix}.

PFMs are therefore also exciting as seeds of a much broader range of novel computer hardware and software opportunities, expanding both the capabilities as well as the physical domain of AI and computer science. While designing large-scale PFMs is undoubtedly very challenging, they are still significantly easier to realize than what one might consider the ultimate machine-learning hardware: hardware that, like our brains, performs not just inference optimized at the absolute lowest physical level, but also learning itself~\cite{momeni2025training}. While such self-learning (and/or self-assembling) hardware could have potential even vastly beyond PFMs, we think that PFMs represent a relatively manageable and economically self-justifying step towards this hardware. By sacrificing programmability, they may allow more efficient (and more scalable) implementation than externally programmable or even self-programmable hardware. Realizing PFMs will likely require large-scale physical inverse design~\cite{molesky2018inverse,sengupta2025ai} that extends modern electronic design automation to the lowest level of hardware abstraction possible. While developing such inverse-design capabilities for hardware at the scale of $>10^{12}$ parameters is clearly a grand challenge, the effort to do so would yield opportunities even at more modest scale (e.g., the inverse design of subparts of a processor, or of analog devices for other purposes besides neural-network inference). The direction PFMs represent is radical, but still largely an evolution of current paradigms: it combines the automated design of algorithms (i.e., machine learning) with automated design of abstracted hardware (i.e., electronic design automation), and extends both, jointly, to the lowest accessible level of hardware physics.

Finally, how large of AI models can be realized with PFMs, and how efficiently could PFMs perform their calculations? While our back-of-the-envelope calculations are uncertain, they suggest PFMs could allow scaling of model sizes by many orders of magnitude beyond current ones, comparable to at least the difference between GPT-2 ($\approx 10^9$ parameters) and GPT-4 ($\approx 10^{12}$ parameters), to $\sim10^{15}$ parameters---and with capabilities that would allow for implementation edge devices. For larger devices that would need remote access and more speculative manufacturing, far greater scaling is possible, up to $10^{18}$, and perhaps even $10^{21}$ parameters. PFMs would also allow for scaling AI systems in other ways. Methods that improve AI system performance by using the model more times during inference (e.g., test-time scaling techniques \cite{cobbe2021training,epoch2023tradingoffcomputeintrainingandinference,snell2024scaling,balachandran2025inference,wu2024inference}) are arguably even more effectively enhanced by PFMs, since PFMs would make inference of larger models dramatically less expensive. \hlredbf{Here, our simplified calculations (Table~\ref{tab:pfm_scaling}) suggest PFMs could reduce energy per inference by $10^4$ to $10^8$ times relative to modern programmable digital processors, and inference latency by roughly $10^3$ to $10^5$ times---with the energy advantage growing, and the speed advantage shrinking, as model size increases.} By enabling such low-cost inference to be implemented in edge devices like robots, PFMs would also expand the applications of AI systems. For AI systems to become broadly useful, they will likely require both scaling and lower costs---for these challenges, PFMs appear to be an unusually high-potential route.

This article's primary purpose is to emphasize (and ask) the many open questions faced by the PFM approach to scaling AI systems. Amid this uncertainty, we do have at least one clear conclusion: if PFMs can be manufactured, if they then work as they are designed to, and finally, if they can even be designed at all, they could radically advance the capabilities and impact of AI.

\section*{Author contributions}
The concept of PFMs emerged through joint ideas and discussions between the authors over several years within P.L.M.'s research group at Cornell University. It was first put into the basic form presented here by L.G.W and T.O., and then with P.L.M. the idea for this article was conceived. Calculations were performed by L.G.W., and figures were created by T.W.. L.G.W and P.L.M. wrote the manuscript.

\section*{Acknowledgements}

We gratefully acknowledge financial support from NTT Research, a David and Lucile Packard Foundation Fellowship (P.L.M.), an Eric and Wendy Schmidt AI in Science Postdoctoral Fellowship (T.W.), the National Science Foundation (Award No. CCF-1918549), the Air Force Office of Scientific Research (Award No. FA9550-22-1-0378 and Award No. FA9550-24-1-0193), and the Department of Energy Office of Science (Award No. DE-SC0026235). We also are grateful for helpful comments and suggestions from Maxwell~Anderson, Vadim~Gutnik, Jérémie~Laydevant, Guilherme~Migliato~Marega, Rishabh~Sehgal, Nikolas~Tezak, Yongqi~Zhang, and Ruomin~Zhu.

\bibliographystyle{mcmahonlab}
\bibliography{references}

\appendix
\section{Additional details on our back-of-the-envelope calculations and platform-specific details for example PFMs}

In this section, we elaborate on several details regarding the back-of-the-envelope calculations presented in the main text.

\subsection{The mathematical analogy between PFM physics and neural network inference}
To ensure that the devices we analyze here are capable of performing calculations of the kind and scale necessary for foundation model inference, it is necessary that we consider how neural-network calculations correspond to the physical processes in each device---that is, we need to assume there exists a mathematical analogy between the calculations we know current artificial neural networks rely on, and the calculations that are natively performed by these physical devices. For simplicity, in each case we consider a kind of loosely implemented multilayer perceptron (MLP) as a representative transformation suitable for foundation-model inference. By ``loosely implemented'', we mean that we have not made an effort to ensure that there is a precise, uncompromising 1:1 mathematical analogy between the physical, analog evolution of the hardware and the mathematical form associated with a multilayer perceptron. We assume that any realistic PFM will require design and modeling that accounts for at least some degree of mismatch between the operations performed natively by the hardware and current artificial neural networks. Nonetheless, as the hardware architectures we have chosen implement \textit{approximately} the same mathematical transformations as an idealized MLP, we are able to make order-of-magnitude estimates of the effective number of parameters and operations performed by the device, which can then be compared (again, to a precision of order of magnitudes) with current digital hardware.

To be clear however, these are assumptions made (only) for the purposes of allowing pedagogical plausibility calculations: physical foundation models will likely need to implement transformations that may be inefficient with even very large multilayer perceptrons, and vice versa, we do not think the most scalable route to construct PFMs is to require that they rely on a human-designed, one-to-one mathematical analogy with existing neural-network architectures. The more promising path, which the calculations of this section merely establish plausibility for, is one of \textit{physical machine learning}, namely the direct machine learning of the physical hardware's design at the lowest possible level of physical abstraction, in essence using the transformations the physical hardware naturally realizes as the mathematical ans\"atze of the physically realized machine-learning model~\cite{wright2022deep,momeni2025training}.

\subsection{Optical Physical Foundation Models}

The possibility of a physical foundation model first occurred to us in the context of photonic computation, where we were inspired by two lines of research: optical data storage, particularly in 3D, and photonic inverse design\footnote{While physical foundation models are a concept whose merit has become clear only with recent progress in artificial intelligence, the opportunity of combining photonic memories and photonic computing is not our original insight. For example, early pioneers in the field of optical neural networks investigated optical computing using holographic optical memory devices based on the photorefractive effect \cite{abu1987optical,caulfield2002optical}, and today many groups pursuing this direction have similarly approached photonic computing hardware by using platforms like phase-change materials that were previously introduced for data storage \cite{rios2019memory,wetzstein2020inference}. Where physical foundation models differ is that we consider the possibility that ``in-memory'' computing may be performed using a fixed, or mostly fixed, memory, rather than a programmable one.}. Optical data storage, such as CD-ROMs, is largely a legacy technology today. However, while optical media are inferior to electronics in their compactness (especially in 2D) due to the much larger wavelength of light, storage of fixed information using optics is nonetheless remarkable---storing petabytes of information in 2D or 3D~\cite{bouwhuis1985principles,sarid2007roadmap,zhang2014seemingly,gu2014optical,zhou2024terabit,zhao20243d,velzel1978laser}, and with non-volatile stability that would allow effectively eternal storage. These works~\cite{bouwhuis1985principles,sarid2007roadmap,zhang2014seemingly,gu2014optical,zhou2024terabit,zhao20243d,velzel1978laser}, spanning several decades and many distinct technologies, show that optics can be used to store a tremendous amount of information stably. To see how this information might be used not as conventionally retrievable data, but instead as the ``weights'' of an extremely complex single-purpose analog computer, we need however to refer to a more recent line of research (at least within the field of photonics\footnote{We note that physical inverse design appeared earlier in mechanical engineering, where it has been applied to the design of a wide range of mechanical parts in, e.g., airplanes, and automobiles \cite{jameson1988aerodynamic,giles2000introduction,martin2004topology}.}), namely photonic inverse design \cite{molesky2018inverse}.

While computer-assisted design has become a ubiquitous part of engineering, physical inverse design as it is commonly practiced in modern photonics research \cite{molesky2018inverse} takes this to its microscopic limit, applying an optimization procedure to microscopic physical simulations to determine a corresponding microscopic design, often without any abstraction. For example, a physical inverse design approach might seek a photonic nanostructure in the form of a pattern of etched silicon on insulator, and optimize the pattern of this etched silicon by performing simulations of Maxwell's equations. To date, the objectives of photonic inverse design have typically been relatively simple, such as realizing a cavity that tightly confines light \cite{albrechtsen2022nanometer}, or creating compact dichroic beamsplitters that route light waves of different colors to different paths \cite{piggott2015inverse}. However, with improved algorithms and computers, larger and larger photonic devices can be inverse-designed, comprising more and more design degrees of freedom. Among many, relevant recent examples are the experimental demonstration of photonic matrix-vector multipliers where the matrix is fixed during fabrication, realized by 3D-printed diffractive plates \cite{lin2018all} and inverse-designed silicon nanostructures \cite{nikkhah2024inverse,zhao2025high}. Photonic inverse design has been primarily applied to realize linear functionalities, but it can be applied also to design nonlinear photonic systems \cite{yanagimoto2025programmable}. Here, relevant examples are inverse-designed nonlinear optical ``neural networks'', wherein the \textit{nonlinear} propagation of optical waves in an inverse-designed nanostructure is used to perform inference of physical computation analogous to a (nonlinear) neural network inference \cite{hughes2019wave,khoram2019nanophotonic,nakajima2021neural}. These works inspired us to ask the following question: If inverse designed devices of this kind continue to scale up in complexity, could we one day realize \textit{photonic} physical foundation models? The box below summarizes our lengthy calculations key findings, which collectively imply that the answer to this question is plausibly: Yes.

\begin{center}
  \setlength{\fboxsep}{8pt}
  \setlength{\fboxrule}{0.8pt}
  \fbox{%
    \begin{minipage}{0.95\linewidth}
      \textbf{The main findings of this calculation.} First, under the assumption of wavelength-scale 3D manufacturing of optical nanostructured media, we find that optical PFMs up to the scale of $10^{18}$ parameters are possible, but would require spectacular fiber-like geometries, e.g, a 50-m long tube with a $5\times5$ cm cross-section. Such a tube would need to be constructed of a medium supporting gain, such as Er- or Yb-doped glass. Second, even assuming that optical nonlinearities due to the Kerr effect are used to allow nonlinear transformations, we find that the energy consumption of optical PFMs will scale with the dimension of the input and output vectors, allowing throughput and energy consumption that improve upon current digital processors by many orders of magnitude.
    \end{minipage}
  }
\end{center}

\noindent\textbf{An example photonic foundation model.} Let us suppose that we want to implement a photonic physical foundation model with $10^{18}$ parameters---this could look like the nanostructured glass shown in Fig. \ref{fig:optical_pfm}. Input vectors to the foundation model would correspond to patterned optical beams, created, for example, using a very high resolution spatial light modulator (SLM). As these beams propagate through the nanostructured glass, the spatial variations of the refractive index within the glass will cause this patterned coherent light to propagate and interfere in complex ways. In the simplest case, where the light is low-intensity and thus propagates linearly through the medium, the transformation of light from the input facet to the output surface is a matrix-vector multiplication \cite{miller2012all}. If higher-intensity light or more strongly nonlinear optical media are used, this operation may become more complex \cite{hughes2019wave,khoram2019nanophotonic,nakajima2021neural}, and could in principle be capable of representing any of the high-dimensional nonlinear transformations realized by modern artificial neural networks. Unless otherwise noted, we will treat this device's operation as a matrix-vector multiplier as a means of estimating its computational throughput; however this is obviously an oversimplification as nonlinear operations are necessary for foundation-model inference, and a conservative lower bound, as nonlinear optical interactions will typically increase the effective number of analog operations performed by the optical wave propagation.

\noindent\textbf{Fitting parameters (into a photonic medium).} As a first-order requirement for a photonic physical foundation model, it needs to be possible to put $10^{18}$ parameters worth of information into a photonic medium, such that all those parameters can be expected to exert a non-negligible influence on the propagation of light through the medium. While light can be affected by strongly subwavelength features, a conservative estimate for the smallest effective feature that may reliably affect an optical wave is one of similar size to the optical wavelength in the medium. If we assume light of wavelength $\lambda=1550$~nm, as in modern telecommunications systems, and a mean refractive index $n$ around $1.5$ (e.g., fused silica glass), then this feature size is about 1000 nm (of course, using shorter wavelengths and/or higher-refractive-index materials could allow even smaller features). The total volume\footnote{We are neglecting here the size of the input laser, the input modulator, and the output detector array, as well as the associated electronics for digital-analog conversion and memory. Input modulator and detector arrays will scale in area as $A\sim N\lambda^2$ and will require substantial circuits for digital-analog conversions, so these could for smaller PFMs provide a bottleneck for compactness. Short-pulse fiber lasers could provide suitable pulses for the requirements we consider here, and would require at least an additional $\sim10$ cm by 10 cm footprint. In short, it is unlikely that photonic PFMs would be competitive as edge devices, though they are promising in the context of remote-accessed inference acceleration.} of a $P$-parameter photonic physical foundation model is then simply $V=P(\frac{\lambda}{n})^3$.

Considering the size required, we think it is most likely that a photonic physical foundation model would be realized in 3D, and would operate on quite high-dimensional input vectors. For example, if we assume an input area of 5 $\times$ 5 cm, we could comfortably accommodate billion-dimensional input vectors\footnote{The number of optical modes that would be guided by such a glass rod, of cross-sectional area $A$, would be about $An^2/\lambda^2$, where $n$ is the typical refractive index and $\lambda$ the vacuum wavelength of the laser light. For $A =$ 5 cm $\times$ 5 cm, $n=1.5$, and $\lambda = 1$ $\mu$m, the number of guided modes is $3.75\times10^{9}$, meaning the rod can support, transversely, slightly more optical spatial modes than we require for $10^9$-dimensional inputs.}. To ensure space for our wavelength-scale parameters, with this cross-sectional area, the length of the photonic computer would need to be 50 m (for $10^{14}$-parameter model, multiply these by $(10^{-4})^{1/3}\approx 0.05$).  In other words, the analog computer would be a long tube or fiber through which light would propagate, realizing a nearly incomprehensibly complex inference calculation passively along the way. To accommodate nonlinearity and the loss of photons by scattering and absorption within this structured material, it will likely be necessary to introduce distributed or periodic optical amplifiers, as in optical telecommunications. (In part as a consequence of this, we think operating the device with laser wavelengths of $\sim$1000~nm, and with more transparent silica glass and mature Yb-based amplifiers, would be a more likely solution than using lossier silicon and 1550 nm lasers.)

\noindent\textbf{Nonlinearity.} For typical laser light, propagation through the structure described so far would produce a linear transformation (i.e., a matrix $M$ multiplying the vector of input optical mode amplitudes), which is insufficient for realizing foundation model inference. In order to ensure that the transformation realized by the photonic physical foundation model is nonlinear, we need the propagation of light through this medium to be optically nonlinear. For example, the Kerr effect is the most common optical nonlinearity, which arises as an optical intensity dependent refractive index perturbation, $n(x,y,z) = n_0(x,y,z) +n_2I(x,y,z)$, where $n_0(x,y,z)$ is the designed 3D refractive index and $n_2$ is the medium's nonlinear refractive index parameter. Accordingly, Kerr nonlinearity means that the matrix realized by propagating through a given photonic medium, $M=M(n(x,y,z))=M(n_0(x,y,z),I(x,y,z))$ is dependent on the input vector, as in the attention mechanism. Of course, the Kerr effect also provides nonlinearity---albeit of a distributed kind that differs from the usual element-wise activation functions of most neural-network architectures.

To access optical nonlinearities, we need a sufficiently high optical intensity such that the Kerr effect gives rise to a non-negligible impact on the propagation of the optical pulse. We can estimate the required pulse intensity by computing the nonlinear phase shift accumulated by a pulse in the medium, neglecting for now the effects of scattering, dispersion, and diffraction. The amount of nonlinear optical phase is $\Phi_{\textrm{NL}}=\frac{2\pi}{\lambda}n_2IL$, where $L$ is the length of the optical medium and $I=P_0/A_0$ is the intensity, equal to the peak power of the laser pulse, $P_0$, divided by its transverse area, $A_0$. For our calculations here, we set $\Phi_{\textrm{NL}}=2\pi$, which corresponds to a moderate level of nonlinearity.

For the optical PFMs described in the main text, and using the nonlinear refractive index of fused silica, $n_2\approx10^{-20}$m$^2$/W, we find that we need a peak power of 44 MW, 450 MW, and 4.5 GW respectively. If we assume 1 ps duration pulses, these correspond to pulse energies of 44 $\mu$J, 440 $\mu$J, and 4.4 mJ (these numbers are rounded up to the nearest factor of 10 in the main text). Experts in nonlinear optics will note that these peak powers are quite high, high enough to exceed the critical power for self-focusing. Normally, sending such intense laser pulses into a long glass tube would result in extremely nonlinear pulse evolution, since the optical beams would collapse to an extremely high intensity spot (or many such spots). However, in the case of the optical PFMs we have reported in the main text, the complex nanostructured refractive index variations would suppress this self-focusing effect \footnote{One can also check that the Marburger self-focusing distance \cite{marburger1975self} is in each case longer than the optical PFM medium, assuming a Gaussian beam of width half the size of the cross-sectional area.} As a consequence, we expect highly nonlinear, but not catastrophically nonlinear, highly complex optical wave evolution.

This type of multimode, highly nonlinear optical wave propagation has been considered before. Multimode nonlinear optical wave propagation has been used as a reservoir computer \cite{teugin2021scalable, ecslik2026multimode, oguz2023forward,oguz2023forward,oguz2024programming,kesgin2025photonic,yuce2025role}, as have works that use nonlinear optical wave propagation as analogues of recurrent neural networks and neural ordinary differential equations \cite{hughes2019wave,khoram2019nanophotonic,nakajima2021neural}. The pattern-formation and nonlinear wave phenomena of nonlinear optical pulse propagation (particularly in amplifying media) resemble those observed with certain neural-field models of biological neural networks \cite{coombes2005waves,ermentrout1998neural}, suggesting that nonlinear wave propagation might appear as an appropriately coarse-grained limit of current neural networks. Moreover, photonic-domain nonlinearity may not even be strictly required, as recent works have shown that linear optical systems may be used to realize nonlinear transformations of data if the input data is encoded into a physical parameter that nonlinearly maps to the light field or its propagation \cite{wanjura2024fully,xia2024nonlinear,yildirim2024nonlinear}, such as a phase shift (e.g., an encoding of data $x$ into the phase of a beam is a nonlinear encoding of that data, $E(x)=Ee^{ix}$). Since linear optical systems are easier to design, obtain repeatable performance from, and allow simpler light sources and materials, such encoding-based nonlinearities stand out as a particularly promising ingredient for photonic PFM architectures.

\noindent\textbf{Input.} We have furthermore assumed an incredible, gigapixel array of modulators and detectors that will likely be very challenging, especially if operated at high speed: Consider that, at 100 MHz, the required data bandwidth for even 2-bit-precision vectors would be 2 bits $\times$ 100 MHz $\times 10^9 \approx$ 25 Petabytes/s. This is higher throughput than current world-record optical communication links (which have reached about 1 Petabit/s \cite{luis20251,jorgensen2022petabit}), so it is a steep requirement.

One way to alleviate this requirement is simply to use lower-dimensional inputs and outputs, and require that these vectors expand into a higher dimensional space within the optical PFM. In this case, high optical pulse energies are still required, but the input and output complexity would be significantly reduced. A second route is simply to tolerate significantly slower inference throughput than 100 MHz. If we opt to restrict the average optical power flowing through the optical PFM to 1 kW (a high value, but still commonly reached in fiber amplifiers), then the inference rate would be limited to 10 MHz, 1 MHz, and 100 kHz. These are significantly longer than the propagation delay, and hence are included in the main text Table.

\subsection{Photonic design optimization and open questions}

While the preceding calculation establishes the attractive possibility of photonic physical foundation models, it also raises many more questions than it resolves. Clearly, the optical PFM we have proposed has many features that would make it challenging to realize. For example, nanostructured 3D (or, perhaps, 2D) optical medium is likely the most physically compact setting for a photonic PFM, but is it the most manufacturable? Does the physical architecture we describe correspond to a mathematical architecture that scales well when trained using gradient descent? In short, is this the best photonic PFM architecture and if not, what is? By ``PFM architecture'' here, we mean a physical structure, and a corresponding mathematical model of it, that specifies the position, range, and effect of adjustable physical parameters and the input and output vectors from the photonic device. An architecture with a set of parameters specifies both the physical device that must be manufactured, but also the (approximate) mathematical transformation that device realizes, i.e., a function of the form $\vec{y}=f_{\textrm{p}}(\vec{x},\vec{\theta})$, where $\vec{x}$ and $\vec{y}$ are the input and output vectors, and $\vec{\theta}$ are the physical parameters, such as the values of the refractive index in the 3D material considered above, $n(x,y,z)$.

Pursuing photonic PFMs will be most effective if this joint physical-mathematical architecture optimization is approached in a progressively more photonics-friendly manner, purposefully using the operations natively performed by manufacturable photonics rather than enforcing the same used in GPU-based deep learning. It may help however to start this process from the familiar initial conditions of established current models, such as Transformers, recurrent neural networks, multilayer perceptrons, and so on, and to then make best-effort attempts to design efficient photonic structures that can approximate these models in manufacturable substrates. These models, particularly when they use very wide layers, are already well-suited to optical acceleration \cite{anderson2024optical}. For example, printed metasurfaces \cite{mcmahon2023physics,wirth2024compressed} or nanostructured silicon-on-insulator waveguides \cite{nikkhah2024inverse,zhao2025high} provide the capacity to realize relatively large matrix-vector multiplications (at least up to vector dimension $N\sim 10^4$), and may be fabricated using standard planar methods. Nonlinearities and operations that depend on the input vector (such as normalization or Attention operations) could initially be offloaded to digital processors \cite{anderson2024optical}. While this would impose a severe bottleneck to the performance of such near-term PFMs, there are likely strategies to scale these models so that they leverage fixed-fabricated photonic compute more and digital co-processing less. Eventually, however, we would need to discover either wholly photonics-aware alternatives to these operations, or to develop photonic PFMs that can approximate them.

\noindent\textbf{Compactness.} An important open question is how compact photonic PFMs may be: it remains plausible that photonic PFMs could be significantly more compact than we have assumed above. Recent results have shown that the required area for implementing an $N\times N$ unitary optical matrix-vector multiplication in 2D devices may not necessarily scale as $A\sim N^2$, as we implicitly assumed above \cite{hamerly2024towards,onodera2024scaling}. Light may be meaningfully affected by subwavelength features, and does not, as we have assumed above, need to propagate solely in one direction through a medium: it may resonate within effective cavities, or undergo strong and multiple scattering events that change its direction, including reversing it entirely. The use of wavelength multiplexing could also allow photonic PFMs to operate on effectively much larger vectors within the same volume. Given that the bandwidth of optics is a key advantage it offers over electronics \cite{mcmahon2023physics}, it seems inevitable that the optimal photonic PFM architecture should be one that, unlike the example we have worked through above, fully exploits this bandwidth.

\noindent\textbf{Other applications.} Finally, a key factor that will influence optimal photonic PFM architectures is the many other applications these devices could, like existing programmable optical neural network accelerators \cite{wetzstein2020inference}, be deployed for. Devices similar to photonic PFMs could provide complex functionalities for accelerating other types of computation beyond large AI model inference, but could also provide rich computational functions in applications within which photonics is already widely employed, such as sensing, imaging, displays, and scientific instruments. Many of these applications do not require multiple functionalities, so a single or limited-purpose optical processor could be effective. While programmable photonic processors \cite{bogaerts2020programmable,perez2025large} are also being intensively researched and developed, they face much harsher challenges than fixed-fabricated photonic devices in terms of speed, size, power consumption and performance, so these two families of photonic processors could prove complementary, both across different applications and within the same device. In short, large-scale photonic inverse design and programmable processors are already being pursued in both industry and academia, and for many good reasons. Photonic PFMs add motivation to push these efforts further.

\end{document}